\documentclass{article}

\usepackage{graphicx}

\usepackage{subcaption}
\usepackage{appendix}

\newcommand{\squeezeup}{\vspace{-2mm}}

\usepackage{hyperref}

\usepackage[accepted]{sysml2019}

\sysmltitlerunning{AI Fairness 360}

\begin{document}

\twocolumn[
\sysmltitle{AI Fairness 360:  An Extensible Toolkit for Detecting, Understanding, and Mitigating Unwanted Algorithmic Bias}



\sysmlsetsymbol{equal}{*}

\begin{sysmlauthorlist}
\sysmlauthor{Rachel K. E. Bellamy}{wrc}
\sysmlauthor{Kuntal Dey}{irld}
\sysmlauthor{Michael Hind}{wrc}
\sysmlauthor{Samuel C. Hoffman}{wrc}
\sysmlauthor{Stephanie Houde}{wrc}
\sysmlauthor{Kalapriya Kannan}{irl}
\sysmlauthor{Pranay Lohia}{irl}
\sysmlauthor{Jacquelyn Martino}{wrc}
\sysmlauthor{Sameep Mehta}{irl}
\sysmlauthor{Aleksandra Mojsilovic}{wrc}
\sysmlauthor{Seema Nagar}{irl}
\sysmlauthor{Karthikeyan Natesan Ramamurthy}{wrc}
\sysmlauthor{John Richards}{wrc}
\sysmlauthor{Diptikalyan Saha}{irl}
\sysmlauthor{Prasanna Sattigeri}{wrc}
\sysmlauthor{Moninder Singh}{wrc}
\sysmlauthor{Kush R. Varshney}{wrc}
\sysmlauthor{Yunfeng Zhang}{wrc}
\end{sysmlauthorlist}

\sysmlaffiliation{wrc}{IBM Research, Yorktown Heights, NY, USA}
\sysmlaffiliation{irl}{IBM Research, Bangalore, India}
\sysmlaffiliation{irld}{IBM Research, New Delhi, India}


\sysmlkeywords{AI Fairness, AI Ethics, AI Bias}

\vskip 0.3in

\begin{abstract}
Fairness is an increasingly important concern as machine learning models are used to support decision making in high-stakes applications such as mortgage lending, hiring, and prison sentencing.  
This paper introduces a new open source Python toolkit for algorithmic fairness, AI Fairness 360 (AIF360), released under an Apache v2.0 license (\url{https://github.com/ibm/aif360}). 
The main objectives of this toolkit are to help facilitate
the transition of fairness research algorithms to use in an industrial setting
and to provide a common framework
for fairness researchers to share and evaluate algorithms.

The package includes a comprehensive set of fairness metrics for datasets and models, explanations for these metrics, and algorithms to mitigate bias in datasets and models.
It also includes an interactive Web experience
(\url{https://aif360.mybluemix.net}) 
that provides a gentle introduction to the concepts and capabilities for line-of-business users,
as well as extensive documentation, usage guidance, and industry-specific tutorials to enable data scientists and practitioners to incorporate the most appropriate tool for their problem into their work products. The architecture of the package has been engineered to conform to a standard paradigm used in data science, thereby further improving usability for practitioners.  Such architectural design and abstractions enable
researchers and developers to extend the toolkit with their new algorithms and improvements, and to use it for performance benchmarking. A built-in testing infrastructure maintains code quality.

\end{abstract}
]



\printAffiliationsAndNotice{}  

\section{Introduction}

Recent years have seen an outpouring of research on fairness and bias in machine learning models. This is not surprising, as fairness is a complex and multi-faceted concept that depends on context and culture.  Narayanan described at least 21 mathematical definitions of fairness from the literature~\cite{Narayanan2018}.
These are not just theoretical differences in how to measure fairness; different definitions produce entirely different outcomes. For example, ProPublica and Northpointe had a public debate on an important social justice issue (recidivism prediction) that was fundamentally about what is the right fairness metric \citep{Larson2016, Dieterich2016,Larson2016a}. Furthermore,  researchers have shown that it is impossible to satisfy all definitions of fairness at the same time \cite{kleinberg2017fairness}.  Thus, although fairness research is a very active field, clarity on which bias metrics and bias mitigation strategies are best is yet to be achieved~\cite{Freidler2018}.

In addition to the multitude of fairness definitions, different bias handling algorithms address different parts of the model life-cycle, and understanding each research contribution, how, when and why to use it is challenging even for experts in algorithmic fairness. As a result, general public, fairness scientific community and AI practitioners need clarity on how to proceed. Currently the burden is on ML and AI developers, as they need to deal with questions such as ``Should the data be debiased?'', ``Should we create new classifiers that learn unbiased models?'', and ``Is it better to correct predictions from the model?'' 

To address these issues we have created AI Fairness 360 (AIF360),
an extensible open source toolkit for detecting, understanding, and mitigating algorithmic biases. The goals of AIF360 are  to promote a deeper understanding of fairness metrics and mitigation techniques; to enable an open common platform for fairness researchers and industry practitioners to share and benchmark their algorithms; and to help facilitate the transition of fairness research algorithms to use in an industrial setting.


AIF360 will make it easier for developers and practitioners to understand bias metrics and mitigation and to foster further contributions and information sharing. 
To help increase the likelihood that AIF360 will develop into a flourishing open source community, we have designed the system to be extensible, adopted software engineering best practices to maintain code quality, and invested significantly in documentation, demos, and other artifacts.

The initial AIF360 Python package implements techniques from 8 published papers from the broader algorithm fairness community.  This includes over 71 bias detection metrics, 9 bias mitigation algorithms, and a unique extensible metric explanations facility to help consumers of the system understand the meaning of bias detection results.  These techniques can all be called in a standard way, similar to scikit-learn's fit/transform/predict paradigm. In addition, there are several realistic tutorial examples and notebooks showing salient features for industry use that can be quickly adapted by practitioners. 


AIF360 is the first system to bring together in one open source toolkit: bias metrics, bias mitigation algorithms, bias metric explanations, and industrial usability. By integrating these aspects, AIF360 can enable stronger collaboration between AI fairness researchers and practitioners, helping to translate our collective research results to practicing data scientists, data engineers, and developers deploying solutions in a variety of industries.

The contributions of this paper are
\begin{itemize}
    \item an extensible architecture that incorporates dataset representations and algorithms for  bias detection, bias mitigation, and bias metric explainability
    \item an empirical evaluation that demonstrates how AIF360 can be used for scientific comparisons of bias metrics and mitigation algorithms
    \item the design of an interactive web experience to introduce users to bias detection and mitigation techniques

\end{itemize}

This paper is organized as follows. In Section \ref{sec:terminology}, we introduce the basic terminology of bias detection and mitigation. In Section \ref{sec:relatedworks}, we review prior art and other open source libraries and contributions in this area. The overall architecture of the toolkit is outlined in Section \ref{sec:architecture}, while Sections \ref{sec:datasetclass}, \ref{sec:metricsclass}, \ref{sec:explainerclass}, and \ref{sec:algorithmsclass} present details of the underlying dataset, metrics, explainer, and algorithms base classes and abstractions, respectively. In Section \ref{sec:testing}, we review our testing protocols and test suite for maintaining quality code. In Section \ref{sec:eval_algs}, we discuss algorithm evaluation in bias checking and mitigation. In Section \ref{sec:webapp}, we describe the design of the front-end interactive experience, and the design of the back-end service. Concluding remarks and next steps are provided in Section \ref{sec:discussion}.

\section{Terminology}
\label{sec:terminology}

In this section, we briefly define specialized terminology from the field of fairness in machine learning. A \emph{favorable label} is a label whose value corresponds to an outcome that provides an advantage to the recipient. Examples are receiving a loan, being hired for a job, and not being arrested. A \emph{protected attribute} is an attribute that partitions a population into groups that have parity in terms of benefit received.
Examples include race, gender, caste, and religion. Protected attributes are not universal, but are application specific. A \emph{privileged} value of a protected attribute indicates a group that has historically been at a systematic advantage. \emph{Group fairness} is the goal of groups defined by protected attributes receiving similar treatments or outcomes. \emph{Individual fairness} is the goal of similar individuals receiving similar treatments or outcomes.  \emph{Bias} is a systematic error. In the context of fairness, we are concerned with unwanted bias that places privileged groups at a systematic advantage and unprivileged groups at a systematic disadvantage. A \emph{fairness metric} is a quantification of unwanted bias in training data or models. A \emph{bias mitigation algorithm} is a procedure for reducing unwanted bias in training data or models. 

\section{Related Work}
\label{sec:relatedworks}

Several open source libraries have been developed in recent years to provide various levels of functionality in learning fair AI models.  Many of these deal only with bias detection, and provide no techniques for mitigating such bias.  Fairness Measures \cite{ Zehlike2017}, for example, provides several fairness metrics, including difference of means, disparate impact, and odds ratio. A set of datasets is also provided, though some datasets are not in the public domain and need explicit permission from the owners to access/use the data.
Similarly, FairML \cite{ Adebayo2016} provides an auditing tool for predictive models by quantifying the relative effects of various inputs on a model’s predictions. This, in turn, can be used to assess the model’s fairness. 
FairTest \cite{Tramer2017}, on the other hand, approaches the task of detecting biases in a dataset by checking for associations between predicted labels and protected attributes.  The methodology also provides a way to identify regions of the input space where an algorithm might incur unusually high errors. This toolkit also includes a rich catalog of datasets. 
Aequitas \cite{Aequitas2018} is another auditing toolkit for data scientists as well as policy makers; it has a Python library as well as an associated web site where data can be uploaded for bias analysis. It offers several fairness metrics, including demographic or statistical parity and disparate impact, along with a "fairness tree" to help users identify the correct metric to use for their particular situation. Aequitas's license does not allow commercial use.
Finally, Themis \cite{Galhotra2017} is an open source bias toolbox that automatically generates test suites to measure discrimination in decisions made by a predictive system.

A handful of toolkits address both bias detection as well as bias mitigation. Themis-ML \cite{Bantilan2018} is one such repository that provides a few fairness metrics, such as mean difference, as well as some bias mitigation algorithms, such as relabeling \cite{KamiranCalders2012}, additive counterfactually fair estimator \cite{Kusner2017}, and reject option classification \cite{Kamiran2012}. 
The repository contains a subset of the methods described in the paper.
Fairness Comparison \cite{Freidler2018} is one of the more extensive libraries. It includes several bias detection metrics as well as bias mitigation methods, including disparate impact remover \cite{FeldmanFMSV2015}, prejudice remover \cite{kamishima2012fairness}, and two-Naive Bayes \cite{Calders2010}. Written primarily as a test-bed to allow different bias metrics and algorithms to be compared in a consistent way, it also allows the addition of additional algorithms and datasets.

Our work on AIF360 aims to unify these efforts and bring together in one open source toolkit a comprehensive set of bias metrics, bias mitigation algorithms, bias metric explanations, and industrial usability. Another contribution of this work is a rigorous architectural design focused on extensibility, usability, explainability, and ease of benchmarking that goes beyond the existing work. We outline these design aspects in more details in the following sections.

\section{Overarching Paradigm and Architecture}
\label{sec:architecture}

\begin{figure*}
    \centering
    \includegraphics[width=\linewidth]{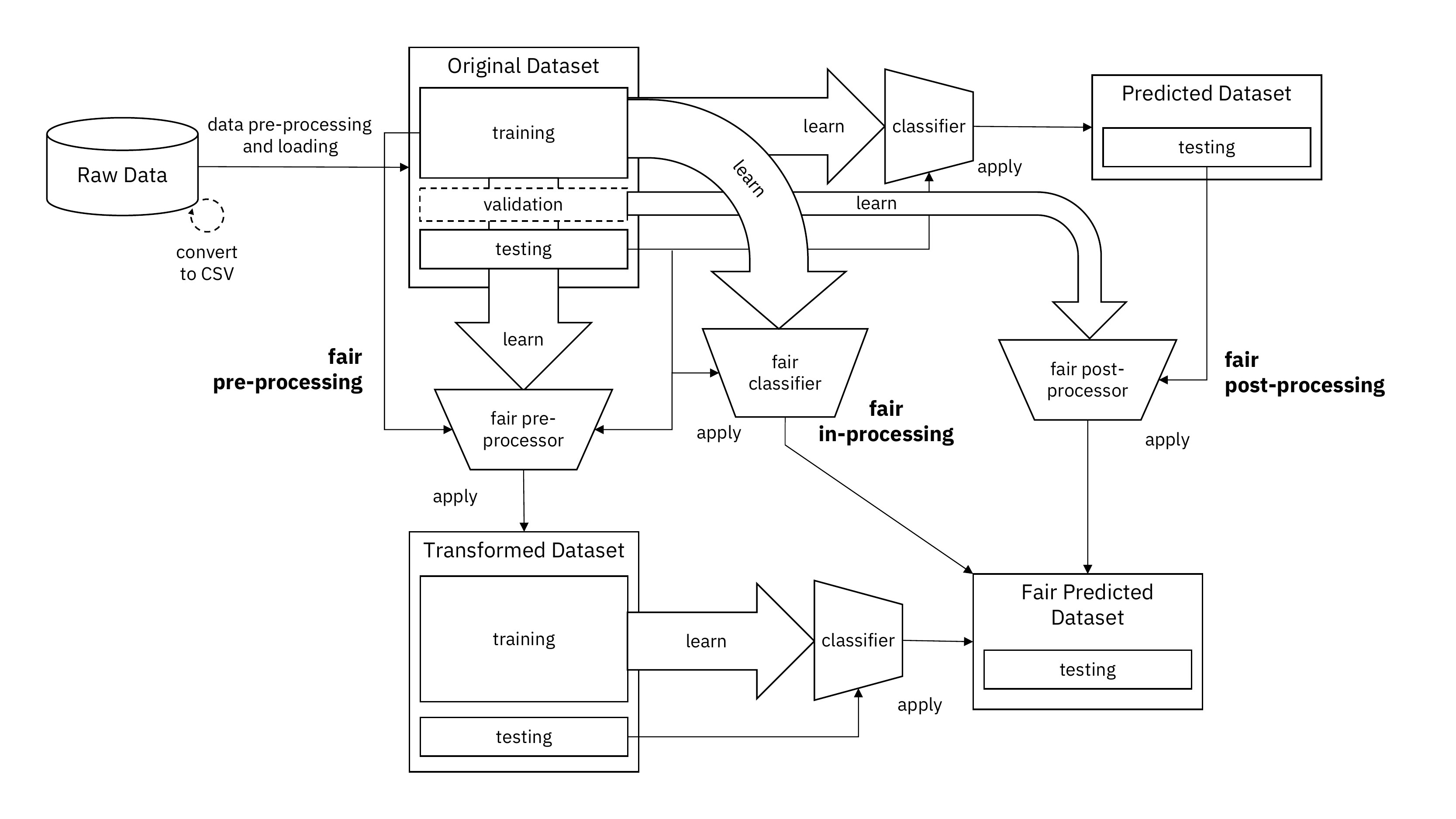}
    \vspace*{-1cm}
    \caption{The fairness pipeline. An example instantiation of this generic pipeline consists of loading data into a dataset object, transforming it into a fairer dataset using a fair pre-processing algorithm, learning a classifier from this transformed dataset, and obtaining predictions from this classifier. Metrics can be calculated on the original, transformed, and predicted datasets as well as between the transformed and predicted datasets. Many other instantiations are also possible.}
    \label{fig:pipeline}
\end{figure*}

AIF360 is designed as an end-to-end workflow with two goals: (a) ease of use and (b) extensibility. Users should be able to go from raw data to a fair model as easily as possible, while comprehending the intermediate results. Researchers should be able to contribute new functionality with minimal effort.

Figure \ref{fig:pipeline} shows our generic pipeline for bias mitigation. Every output in this process (rectangles in the figure) is a new dataset that shares, at least, the same protected attributes as other datasets in the pipeline. Every transition is a transformation that may modify the features or labels or both between its input and output. Trapezoids represent learned models that can be used to make predictions on test data. There are also various stages in the pipeline where we can assess if bias is present using fairness metrics (not pictured), and obtain relevant explanations for the same (not pictured). These will each be discussed below.

To ensure ease of use, we created simple abstractions for datasets, metrics, explainers, and algorithms. Metric classes compute fairness and accuracy metrics using one or two datasets, explainer classes provide explanations for the metrics, and algorithm classes implement bias mitigation algorithms. Each of these abstractions are discussed in detail in the subsequent sections. The term ``dataset'' refers to the dataset object created using our abstraction, as opposed to a CSV data file or a Pandas \texttt{DataFrame}.

The base classes for the abstractions are general enough to be useful, but specific enough to prevent errors. For example, there is no functional difference between predictions and ground-truth data or training and testing data. Each dataset object contains both features and labels, so it can be both an output of one transformation and an input to another. More specialized subclasses benefit from inheritance while also providing some basic error-checking, such as determining what metrics are available for certain types of datasets. Finally, we are able to generate high-quality, informative documentation automatically by using Sphinx\footnote{\url{http://www.sphinx-doc.org/en/master/}} to parse the docstring blocks in the code.

There are three main paths to the goal of making fair predictions (bottom right) --- these are labelled in bold: fair pre-processing, fair in-processing, and fair post-processing. Each corresponds to a category of bias mitigation algorithms we have implemented in AIF360. Functionally, however, all three classes of algorithms act on an input dataset and produce an output dataset. This paradigm and the terminology we use for method names are familiar to the machine learning/data science community and similar to those used in other libraries such as scikit-learn.\footnote{\url{http://scikit-learn.org}} Block arrows marked ``learn'' in Figure \ref{fig:pipeline} correspond to the \texttt{fit} method for a particular algorithm or class of algorithms. Sequences of arrows marked ``apply'' correspond to \texttt{transform} or \texttt{predict} methods. Predictions, by convention, result in an output that differs from the input by labels and not features or protected attributes. Transformations result in an output that may differ in any of those attributes.

Although pre-, in-, and post-processing algorithms are all treated the same in our design, there are important considerations that the user must make in choosing which to use. For example, post-processing algorithms are easy to apply to existing classifiers without retraining. By making the distinction clear, which many libraries listed in Section \ref{sec:relatedworks} do not do, we hope to make the process transparent and easy to understand.

As for extensibility, while we cannot generalize from only one user, we were very encouraged about how easy it is to use the toolkit when within days of the toolkit being made available, a researcher from the AI fairness field submitted a pull request asking to add his group's bias mitigation algorithm. In a subsequent interview, this contributor informed us that contributing to the toolkit did not take much time as: ``\textit{...it was very well structured and very easy to follow}''.  

A simplified UML class diagram of the code is provided in Appendix \ref{appendix:uml} for reference. Code snippets for an instantiation of the pipeline based on our AIF360 implementation is provided in Appendix \ref{appendix:snippets}.



\section{Dataset Class}
\label{sec:datasetclass}

The \texttt{Dataset} class and its subclasses are a key abstraction that handle all forms of data. Training data is used to learn classifiers.  Testing data is used to make predictions and compare metrics. Besides these standard aspects of a machine learning pipeline, fairness applications also require associating protected attributes with each instance or record in the data. To maintain a common format, independent of what algorithm or metric is being applied, we chose to structure the \texttt{Dataset} class so that all of these relevant attributes --- features, labels, protected attributes, and their respective identifiers (names describing each) --- are present and accessible from each instance of the class. Subclasses add further attributes that differentiate the dataset and dictate with which algorithms and metrics it is able to interact.

Structured data is the primary form of dataset studied in the fairness literature and is represented by the \texttt{StructuredDataset} class. Further distinction is made for a \texttt{BinaryLabelDataset} --- a structured dataset that has a single label per instance that can only take one of two values: favorable or unfavorable.
Unstructured data, which is receiving more attention from the fairness community, 
can be accommodated in our architecture by constructing a 
parallel class to the \texttt{StructuredDataset} class, without affecting existing classes or algorithms that are not applicable to unstructured data.

The classes provide a common structure for the rest of the pipeline to use.  However, since raw data comes in many forms, it is not possible to automatically load an arbitrary raw dataset without input from the user about the data format. The toolkit includes a \texttt{StandardDataset} class that standardizes the process of loading a dataset from CSV format and populating the necessary class attributes. Raw data must often be ``cleaned'' before being used and categorical features must be encoded as numerical entities. Furthermore, with the same raw data, different experiments are often run using subsets of features or protected attributes. The \texttt{StandardDataset} class handles these common tasks by providing a simple interface for the user to specify the columns of a Pandas \texttt{DataFrame} that correspond to features, labels, protected attributes, and optionally instance weights; the values of protected attributes that correspond to privileged status; the values of labels that correspond to favorable status; the features that need to be converted from categorical to numerical; and the subset of features to keep for the subsequent analysis.
It also allows for arbitrary, user-defined data pre-processing, such as deriving new features from existing ones or filtering invalid instances.

We extend the \texttt{StandardDataset} class with examples of commonly used datasets that can be used to load datasets in different manners without modifying code by simply passing different arguments to the constructor. This is in contrast with other tools that make it more difficult to configure the loading procedure at runtime. We currently provide an interface to seven popular datasets: \textit{Adult Census Income} \cite{Adult_Dataset}, \textit{German Credit} \cite{German_Dataset}, \textit{ProPublica Recidivism (COMPAS)} \cite{Compas_Dataset}, \textit{Bank Marketing} \cite{Bank_Dataset}, and three versions of \textit{Medical Expenditure Panel Surveys} \cite{MEPS_Dataset_15,MEPS_Dataset_16}.

Besides serving as a structure that holds data to be used by bias mitigation algorithms or metrics, the \texttt{Dataset} class provides many important utility functions and capabilities. Using Python's \texttt{==} operator, we are able to compare equality of two datasets and even compare a subset of fields using a custom context manager \texttt{temporarily\_ignore}\footnote{A sample snippet of this functionality is as follows:

\texttt{with transf.temporarily\_ignore(`labels'):}

\texttt{\hspace{6ex}return transf == pred}

This returns \texttt{True} if the two datasets \texttt{transf} and \texttt{pred} differ only in labels.
} The \texttt{split} method allows for easy partitioning into training, testing, and possibly validation sets. We are also able to easily convert to Pandas \texttt{DataFrame} format for visualization, debugging, and compatibility with externally implemented code. Finally, we track basic metadata associated with each dataset instance. Primary among these is a simple form of provenance-tracking: after each modification by an algorithm, a new object is created and a pointer to the previous state is kept in the metadata along with details of the algorithm applied. This way, we maintain trust and transparency in the pipeline.


\section{Metrics Class}
\label{sec:metricsclass}

The \texttt{Metric} class and its subclasses compute various individual and group fairness metrics to check for bias in datasets and models.\footnote{In the interest of space, we omit mathematical notation and definitions here. They may be found elsewhere, including the documentation for AIF360.}  The \texttt{DatasetMetric} class and its subclass \texttt{BinaryLabelDatasetMetric} examine a single dataset as input (\texttt{StructuredDataset} and \texttt{BinaryLabelDataset}, respectively) and are typically applied in the left half of Figure \ref{fig:pipeline} to either the original dataset or the transformed dataset.  The metrics therein are the group fairness measures of disparate  (DI) and statistical parity difference (SPD) --- the ratio and difference, respectively, of the base rate conditioned on the protected attribute --- and the individual fairness measure \emph{consistency} defined by \citet{zemel2013fairrepresentations}.

In contrast, the \texttt{SampleDistortionMetric} and \texttt{ClassificationMetric} classes examine two datasets as input. For classification metrics, the first input is the original or transformed dataset containing true labels, and the second input is the predicted dataset or fair predicted dataset, respectively, containing predicted labels.  This metric class implements accuracy and fairness metrics on models.  The sample distortion class contains distance computations between the same individual point in the original and transformed datasets for different distances. Such metrics are used, e.g. by \citet{CalmonWVRV2017}, to quantify individual fairness.  

A large collection of group fairness and accuracy metrics in the classification metric class are functions of the confusion matrix of the true labels and predicted labels, e.g. false negative rate difference and false discovery rate ratio. Two metrics important for later reference are average odds difference (the mean of the false positive rate difference and the true positive rate difference), and equal opportunity difference (the true positive rate difference).  Classification metrics also include disparate impact and statistical parity difference, but based on predicted labels.  Since the main computation of confusion matrices is common for a large set of metrics, we utilize memoization and caching of computations for performance on large-scale datasets.  This class also contains metrics based on the generalized entropy index, which is able to quantify individual and group fairness with a single number \cite{Speicher2018}. Additional details on the metrics used in the evaluations (Section \ref{sec:eval_algs}) are given in Appendix \ref{appendix:expt_details}.

There is a  need for a large number and variety of fairness metrics in the toolkit because there is no one best metric relevant for all contexts. It must be chosen carefully, based on subject matter expertise and worldview \citep{FriedlerSV2016}. The comprehensiveness of the toolkit allows a user to not be hamstrung in making the most appropriate choice. 

\section{Explainer Class}
\label{sec:explainerclass}


The \texttt{Explainer} class is intended to be associated with the \texttt{Metric} class and provide further insights about computed fairness metrics.  Different subclasses of varying complexity that extend the \texttt{Explainer} class can be created to output explanations that are meaningful to different user personas.  To the best of our knowledge, this is the first fairness toolkit that stresses the need for explanations.  The explainer capability implemented in the first release of AIF360 is basic reporting through "pretty print" and JSON outputs.  Future releases may include methodologies such as fine-grained localization of bias (we describe the approach herein), actionable recourse analysis \cite{UstunSL2018}, and counterfactual fairness \cite{WachterMR2018}. 



\subsection{Reporting}
\label{subsec:explainer:interface}



\texttt{TextExplainer}, a subclass of \texttt{Explainer}, returns a plain text string with a metric value. For example, the explanation for the accuracy metric is simply the text string \textit{``Classification accuracy on $\langle$count$\rangle$ instances: $\langle$accuracy$\rangle$''}, where $\langle$count$\rangle$ represents the number of records, and $\langle$accuracy$\rangle$ the accuracy. This can be invoked for both the privileged and unprivileged instances by passing arguments.

\texttt{JSONExplainer} extends \texttt{TextExplainer} and produces three output attributes in JSON format: (a) meta-attributes about the metric such as its name, a natural language description of its definition and its ideal value in a bias-free world, (b) statistical attributes that include the raw and derived numbers, and (c) the plain text explanation passed unchanged from the superclass \texttt{TextExplainer}. Outputs from this class are consumed by the Web application described in Section \ref{sec:webapp}.
\subsection{Fine-grained localization}

\begin{figure}
\centering
\begin{tabular}{cc}
\includegraphics[height=1.4in]{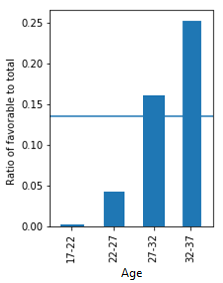} & \includegraphics[height=1.4in]{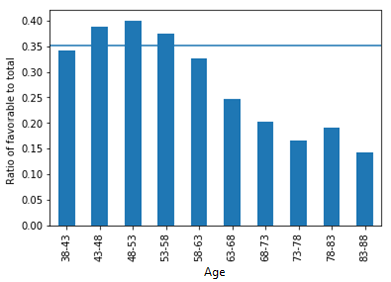}\\
\footnotesize{(a)} & \footnotesize{(b)}
\end{tabular}
\caption{Protected attribute bias localization in (a) younger (unprivileged), and (b) older (privileged) groups in the \textit{German Credit} dataset. The 17--27 range in the younger group and the 43--58 range in the older group would be localized by the approach.}
\label{fig:source_bias_numerical_fairnes_attr}
\end{figure}

\begin{figure}
\centering
    \includegraphics[height=2in]{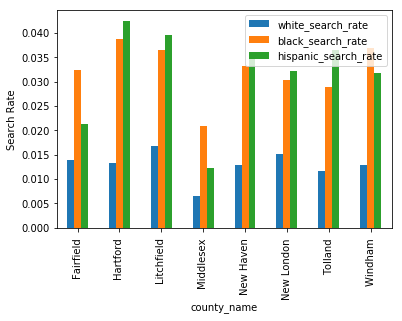}
    \caption{Feature bias localization in the \emph{Stanford Open Policing} dataset for Connecticut, with county name as the feature and race as the protected attribute. In Hartford County, the ratio of search rates for the unprivileged groups (black and Hispanic) in proportion to the search rate for the privileged group (this ratio is the DI fairness metric) is higher than the same metric in Middlesex County and others. The approach would localize Hartford County.}
  \label{fig:CT_search_rate}
\end{figure}

A more insightful explanation for fairness metrics is the localization of the source of bias at a fine granularity in the protected attribute and feature spaces.  In the protected attribute space, the approach finds the values in which the given fairness metric is diminished (unprivileged group) or enhanced (privileged group) compared to the entire group. 
In the feature space, the approach computes the given fairness metric across all feature values and localizes on ones that are most objectionable.  Figure \ref{fig:source_bias_numerical_fairnes_attr} illustrates protected attribute bias localization on the \emph{German Credit} dataset, with age as the protected attribute. Figure \ref{fig:CT_search_rate} illustrates feature bias localization on the \emph{Stanford Open Policing} dataset \citep{pierson2017large} for Connecticut, with county name as the feature and race as the protected attribute.

\section{Algorithms Class}
\label{sec:algorithmsclass}

Bias mitigation algorithms attempt to improve the fairness metrics by modifying the training data, the learning algorithm, or the predictions. These algorithm categories are known as pre-processing, in-processing, and post-processing, respectively \citep{DalessandroOL2017}.

\subsection{Bias mitigation approaches}

The bias mitigation algorithm categories are based on the location where these algorithms can intervene in a complete machine learning pipeline. If the algorithm is allowed to modify the training data, then pre-processing can be used. If it is allowed to change the learning procedure for a machine learning model, then in-processing can be used. If the algorithm can only treat the learned model as a black box without any ability to modify the training data or learning algorithm, then only post-processing can be used. This is illustrated in Figure \ref{fig:pipeline}.

\subsection{Algorithms}

AIF360 currently contains $9$ bias mitigation algorithms that span these three categories. All the algorithms are implemented by inheriting from the \texttt{Transformer} class. Transformers are an abstraction for any process that acts on an instance of \texttt{Dataset} class and returns a new, modified \texttt{Dataset} object. This definition encompasses pre-processing, in-processing, and post-processing algorithms.

\noindent{\textbf{Pre-processing algorithms:}} Reweighing \citep{KamiranCalders2012} generates weights for the training examples in each (group, label) combination differently to ensure fairness before classification. Optimized preprocessing \citep{CalmonWVRV2017} learns a probabilistic transformation that edits the features and labels in the data with group fairness, individual distortion, and data fidelity constraints and objectives. Learning fair representations \citep{zemel2013fairrepresentations} finds a latent representation that encodes the data well but obfuscates information about protected attributes. Disparate impact remover \citep{FeldmanFMSV2015} edits feature values to increase group fairness while preserving rank-ordering within groups.

\noindent{\textbf{In-processing algorithms:}} Adversarial debiasing \citep{ZhangLM2018} learns a classifier to maximize prediction accuracy and simultaneously reduce an adversary’s ability to determine the protected attribute from the predictions. This approach leads to a fair classifier as the predictions cannot carry any group discrimination information that the adversary can exploit. Prejudice remover \citep{kamishima2012fairness} adds a discrimination-aware regularization term to the learning objective.

\noindent{\textbf{Post-processing algorithms:}} Equalized odds postprocessing \citep{hardt2016equality} solves a linear program to find probabilities with which to change output labels to optimize equalized odds. Calibrated equalized odds postprocessing \citep{pleiss2017fairness} optimizes over calibrated classifier score outputs to find probabilities with which to change output labels with an equalized odds objective. Reject option classification \citep{Kamiran2012} gives favorable outcomes to unprivileged groups and unfavorable outcomes to privileged groups in a confidence band around the decision boundary with the highest uncertainty.

\section{Maintaining Code Quality}
\label{sec:testing}

Establishing and maintaining high quality code is crucial for an evolving open source system. Although we do not claim any novelty over modern software projects, we do feel that faithfully adopting such practices is another distinguishing feature of AIF360 relative to other fairness projects.

An extensible toolkit should provide  confidence to its contributors that, while it makes it is easy for them to extend, it does not alter the existing API contract. Our testing protocols are designed to cover software engineering aspects and comprehensive test suites that focus on the performance metrics of fairness detection and mitigation algorithms. 

The AIF360 Github repository is directly integrated with Travis CI,\footnote{\url{https://travis-ci.org/}} a continuous testing and integration framework, which invokes {\tt pytest} to run our unit tests. Any pull request automatically triggers the tests.  The results of the tests are made available to the reviewer of the pull request to help ensure that changes to the code base do not introduce bugs that would break the tests.

Unit test cases ensure that classes and functions defined in the different libraries are functionally correct and do not break the flow of the fairness detection and mitigation pipeline.  Each of our classes is equipped with unit tests that attempt to cover every aspect of the class/module/functions.  

We have also developed a test suite to compute the metrics reported in Section \ref{sec:metricsclass}. Our measurements include aspects of the fairness metrics, classification metrics, dataset metrics, and distortion metrics, covering a total of 71 metrics at the time of this writing. These metrics tests can be invoked directly with any fairness algorithm.
The test suite also provides unit tests for all bias mitigation algorithms 
and basic validation of  the datasets.   
 
 Our repository has two types of tests: (1) unit tests that test individual helper functions, 
and (2) integration tests that test a complete flowof bias mitigation algorithms in Jupyter notebooks. Table \ref{tab:devopstats} provides the statistics and code coverage information as reported by the tool \texttt{py.test --cov} and Jupyter notebook coverage using \texttt{py.test --nbval} .

 \begin{table}[h!]
 \begin{scriptsize}
 \caption{Statistics on the Test Suite for AIF360}
     \label{tab:devopstats}
     \begin{center}
     \begin{tabular}{|c|c|}
     \hline
      Metrics & Statistics \\
      \hline
         Number of Unit Test cases & 23 test cases in 13 modules \\
          Code Coverage (Helper Files) & 65\%\\
          Code Coverage (Algorithms) & 58\%\\
    \hline
     \end{tabular}
     \end{center}
 \end{scriptsize}
 \end{table}
 \squeezeup

\section{Evaluation of the Algorithms}


\label{sec:eval_algs}
Fairness is a complex construct that cannot be captured with a one-size-fits-all solution. Hence, our goal in this evaluation is two-fold: (a) demonstrating the capabilities of our toolkit in terms of the various fairness metrics and bias mitigation algorithms, (b) showing how a user can understand the behavior of various metrics and bias mitigation algorithms on her dataset, and make an appropriate choice according to her needs. 

 \begin{table} [h!]
 \begin{scriptsize}
     \caption{Overview of the experimental setup}
     \label{tab:evaluation}
     \centering
     \begin{tabular}{|c|c|}
     \hline
      Datasets & Adult Census Income, German Credit, COMPAS \\
      \hline
      Metrics & Disparate impact\\
              & Statistical parity difference \\
              & Average odds difference \\
              & Equal opportunity difference\\
     \hline
     Classifiers & Logistic regression (LR),\\
                 & Random forest classifier (RF), Neural Network (NN)\\
     \hline
       & Re-weighing \cite{KamiranCalders2012} \\
      Pre-processing & Optimized pre-processing \cite{CalmonWVRV2017} \\
       Algorithms & Learning fair representations \cite{zemel2013fairrepresentations}\\
                               & Disparate impact remover \cite{FeldmanFMSV2015} \\
     \hline
     In-processing & Adversarial debiasing \cite{ZhangLM2018} \\
     Algorithms   & Prejudice remover \cite{kamishima2012fairness} \\
     \hline
     Post-processing & Equalized odds post-processing \cite{hardt2016equality} \\
    Algorithms & Calibrated eq. odds postprocessing \cite{pleiss2017fairness}\\
    & Reject option classification \cite{Kamiran2012}\\
    \hline
     \end{tabular}
\end{scriptsize}
 \end{table}
 
Table \ref{tab:evaluation} provides the datasets, metrics, classifiers, and bias mitigation algorithms used in our experiments. Additional details on the datasets and metrics are available in Appendix \ref{appendix:expt_details}. The processed \textit{Adult Census Income}, \textit{German Credit}, and \textit{COMPAS} datasets contain $45{,}222$, $1{,}000$ and $6{,}167$ records respectively. Except Adversarial debiasing and Disparate impact remover, all other bias mitigation algorithms use datasets that are cleaned and pre-processed in a similar way. Each dataset is randomly divided into $50\%$ training, $20\%$ validation, and $30\%$ test partitions. Each point in the figures of results consists of a mean and a spread ($\pm$1  standard deviation) computed using $25$ such random splits. For the random forest classifier, we set the number of trees to be $100$, and the minimum samples at a leaf node to be $20$.

\begin{figure*}[h]
 \begin{scriptsize}
     \begin{subfigure}{1.0\textwidth}
        \centering
        \includegraphics[scale=0.30]{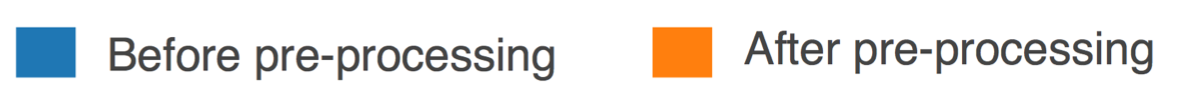}
    \end{subfigure}
    \begin{subfigure}{0.24\textwidth}
      \centering
      \includegraphics[ width=1.7in]{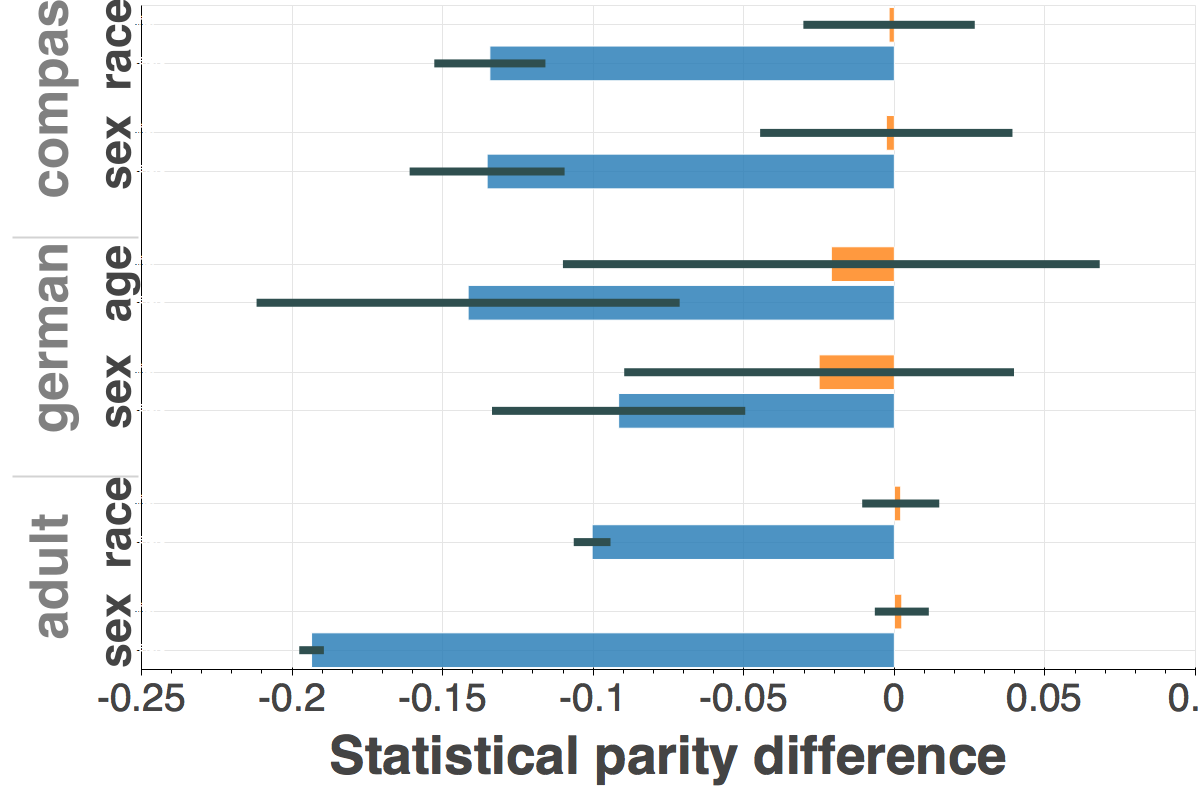}
    \caption{SPD - Re-weighing}
    \label{fig:reweighing_statistical_parity_difference}
    \end{subfigure}
    \begin{subfigure}{0.24\textwidth}
      \centering
     \includegraphics[ width=1.7in]{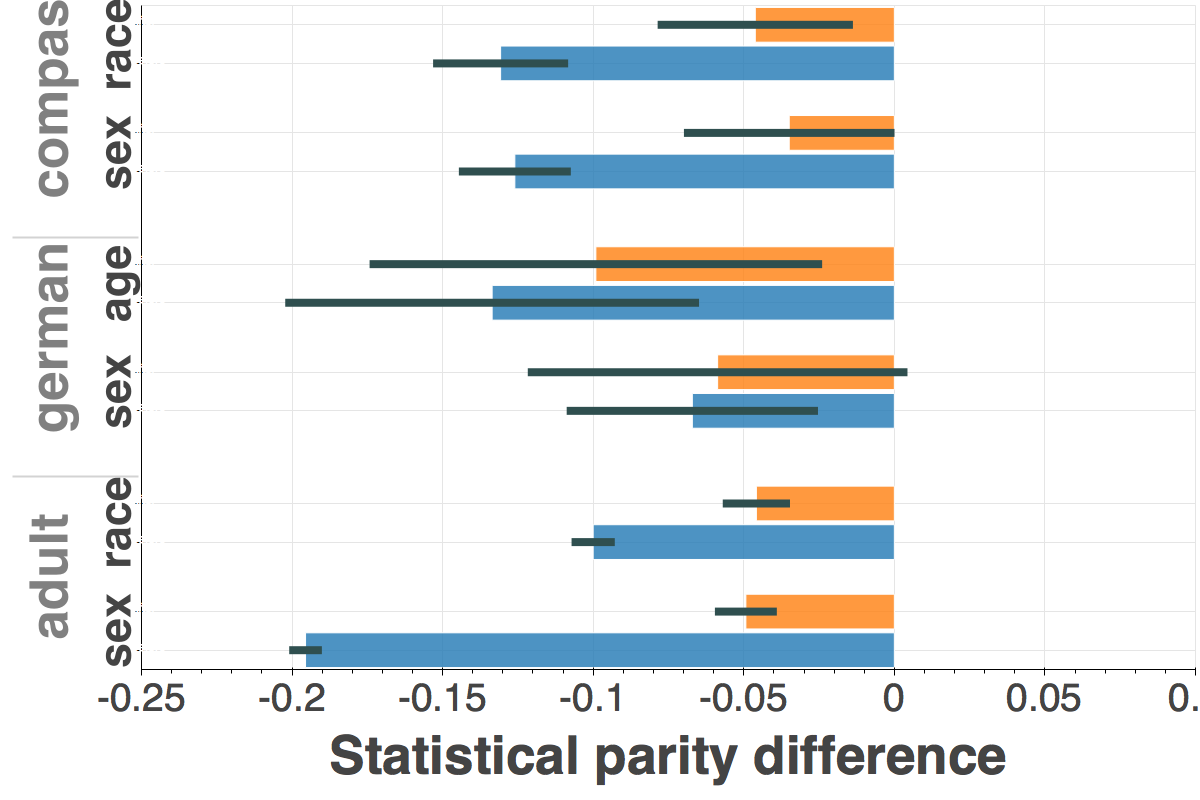}
    \caption{SPD - Optimized pre-proc.}
    \label{fig:optim_preproc_statistical_parity_difference}
    \end{subfigure}
    \begin{subfigure}{0.24\textwidth}
      \centering
      \includegraphics[ width=1.7in]{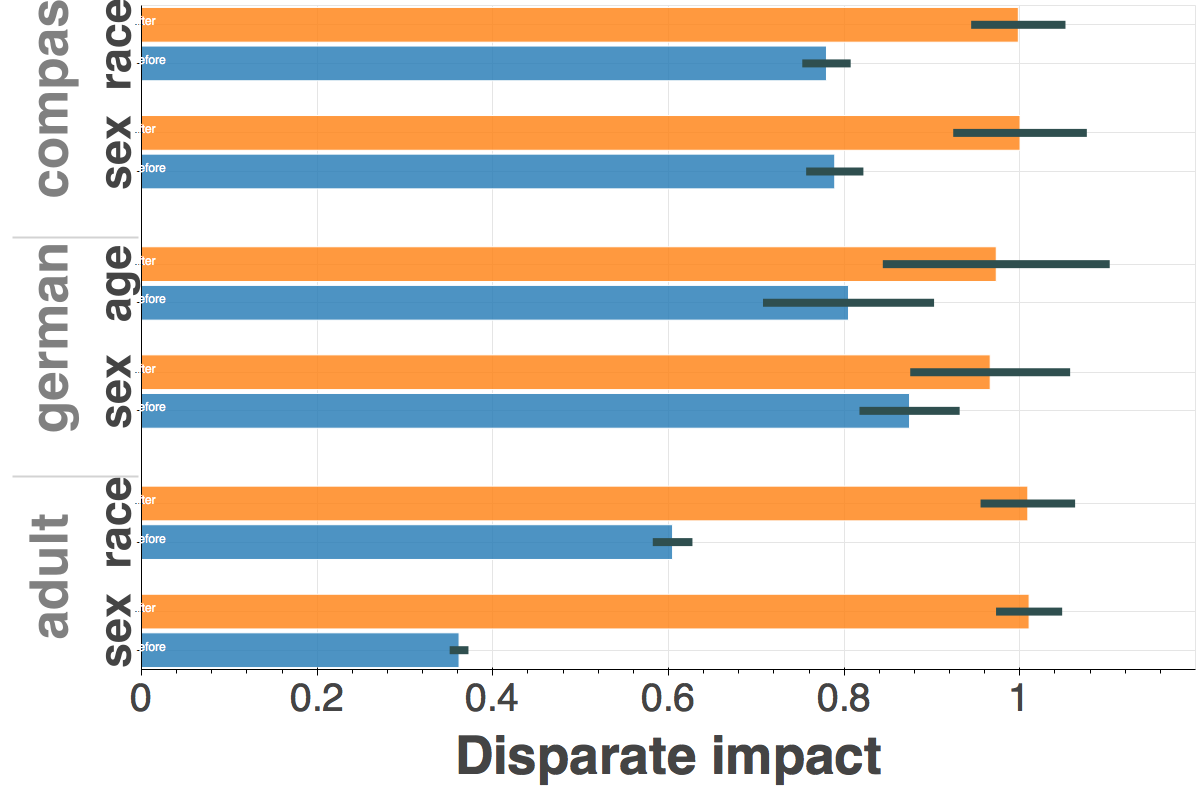}
    \caption{DI - Re-weighing}
    \label{fig:reweighing_disparate_impact}
    \end{subfigure}
    \begin{subfigure}{0.24\textwidth}
      \centering
      \includegraphics[ width=1.7in]{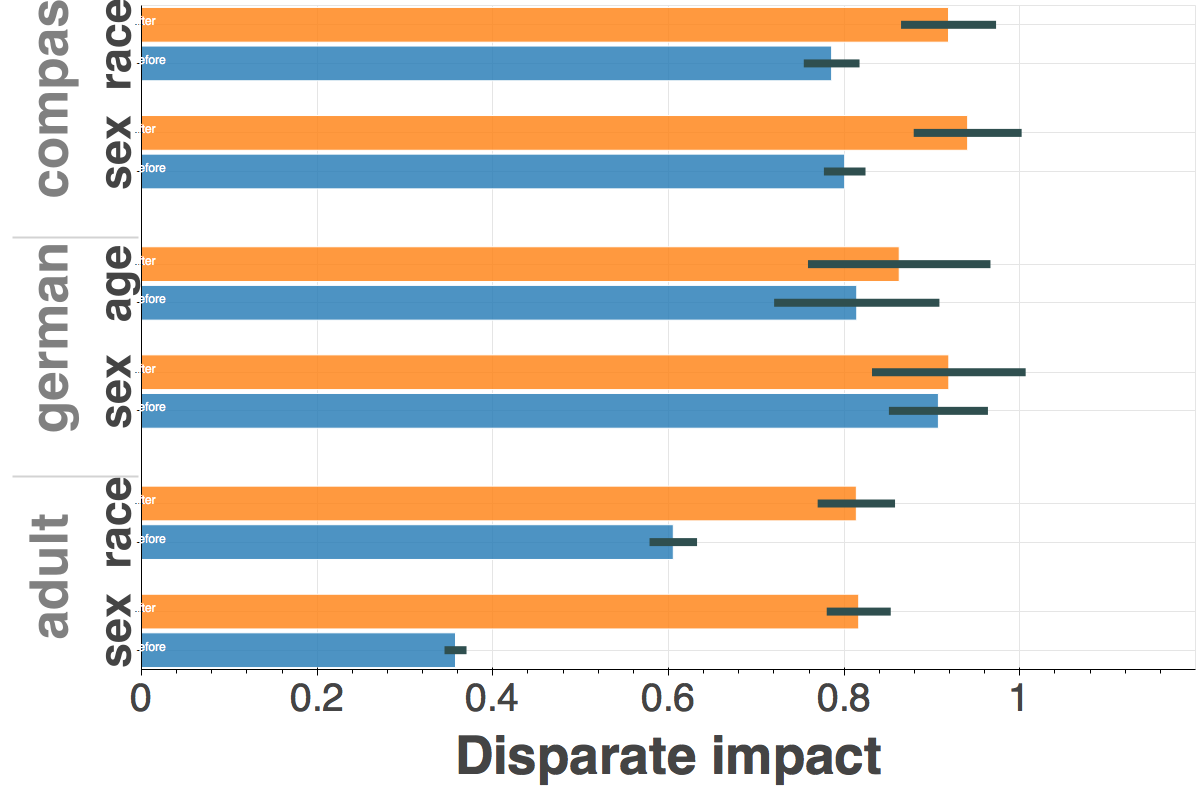}
    \caption{DI - Optimized pre-proc.}
    \label{fig:optim_preproc_disparate_impact}
    \end{subfigure}
\caption{Statistical Parity Difference (SPD) and Disparate Impact (DI) before (blue bar) and after (orange bar) applying pre-processing algorithms on various datasets for different protected attributes. The dark gray bars indicate the extent of $\pm$1 standard deviation. The ideal fair value of SPD is 0 and DI is 1.}
\label{fig:spddi}
 \end{scriptsize}
\end{figure*}

For fair pre-processing algorithms, since the original dataset itself gets transformed (see Figure \ref{fig:pipeline}), we compute fairness metrics before and after this transformation and present results in Figure \ref{fig:spddi}. For all datasets, the Re-weighing and Optimized pre-processing algorithms improve fairness in both metrics presented. However, the least improvement is with \textit{German Credit} dataset, possibly because it is the smallest in size. Results for disparate impact remover and learning fair representations algorithms are not shown since they do not modify the labels or protected attributes directly when transforming the dataset. Hence the SPD and DI values do not change during transformation.




\begin{figure*}
\begin{scriptsize}
\begin{subfigure}{1.0\textwidth}
    \centering
    \includegraphics[scale=0.25]{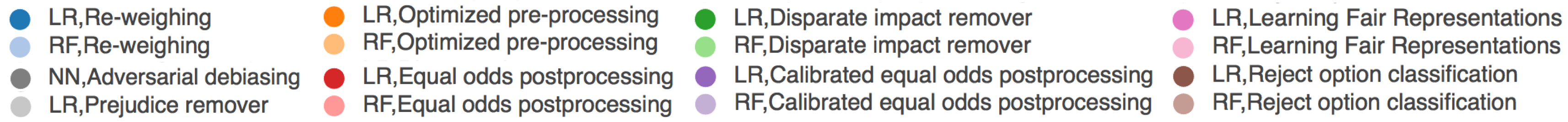}
\end{subfigure}
\begin{subfigure}{0.245\textwidth}
    \centering
    \includegraphics[scale=0.1]{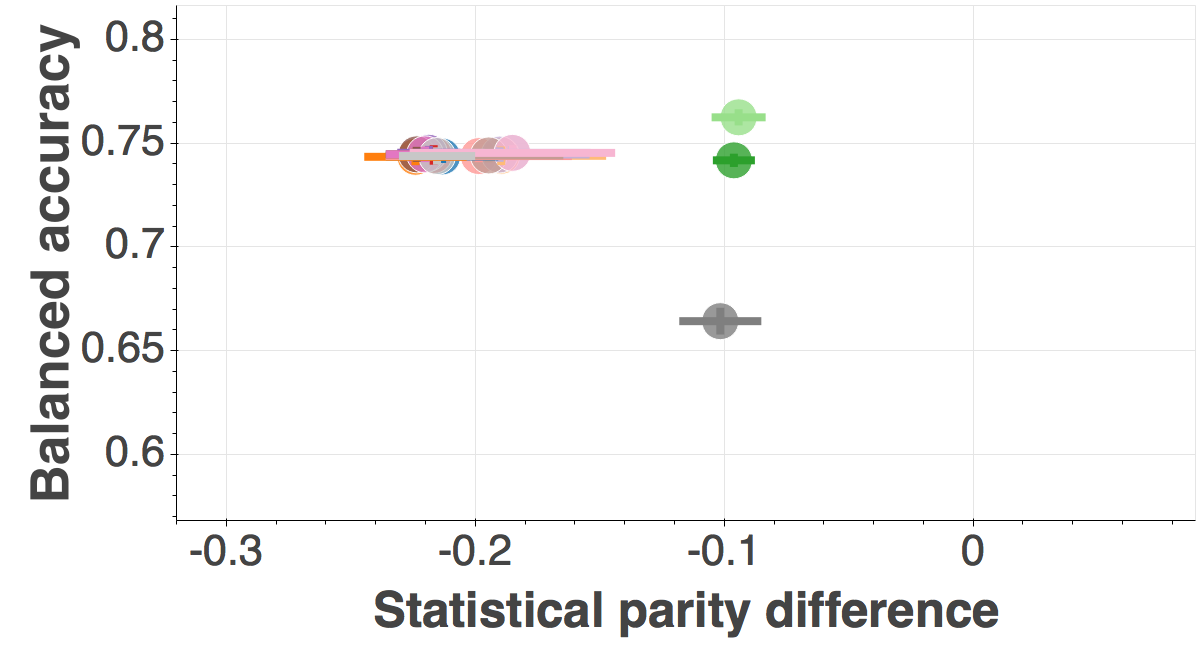}
    \includegraphics[scale=0.1]{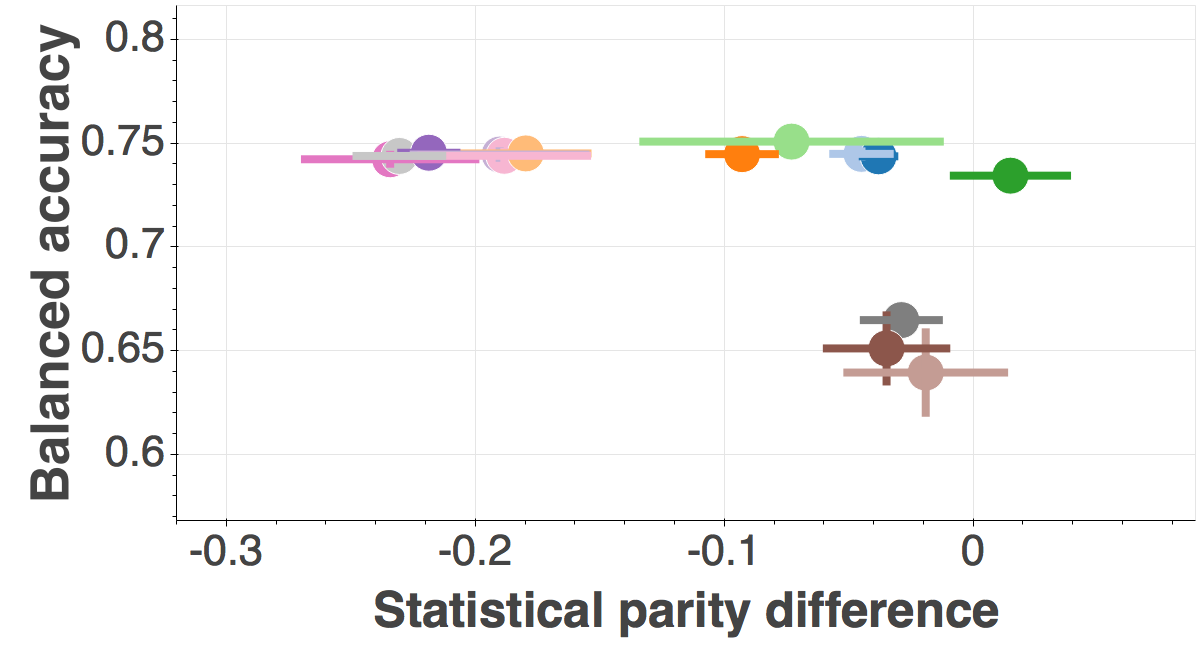}
    \label{fig:adult_race_statistical_parity_difference}
    \caption{Statistical parity difference}
\end{subfigure}
\begin{subfigure}{0.245\textwidth}
    \centering
    \includegraphics[scale=0.1]{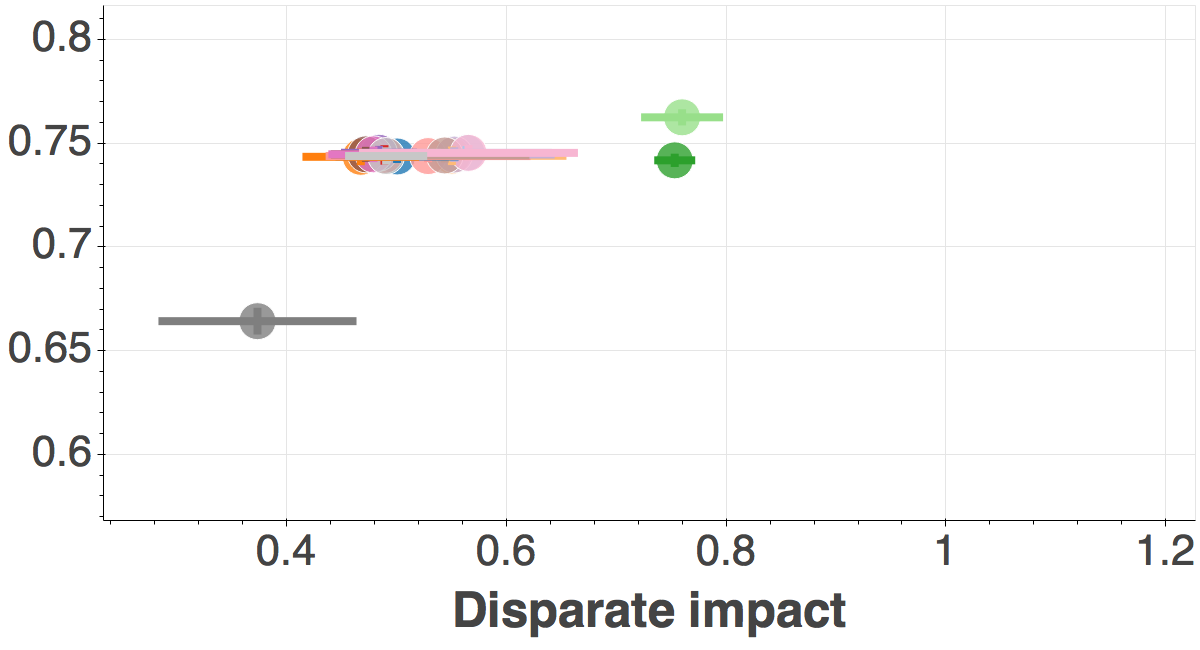}
    \includegraphics[scale=0.1]{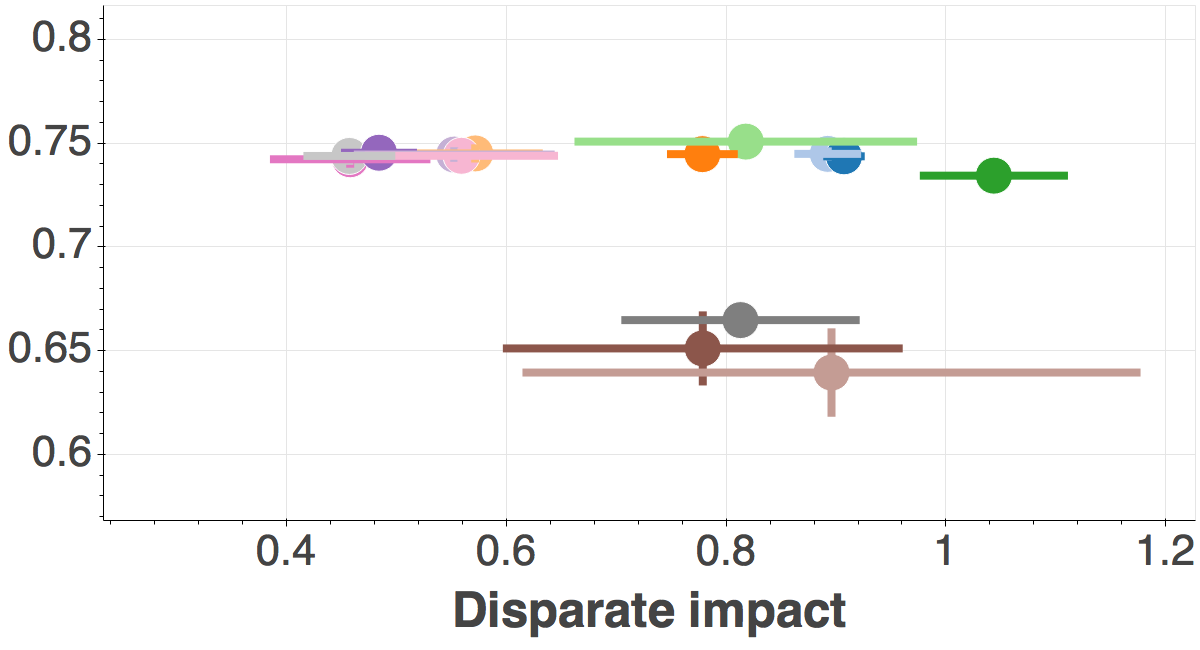}
    \label{fig:adult_race_disparate_impact}
    \caption{Disparate impact}
\end{subfigure}
\begin{subfigure}{0.245\textwidth}
    \centering
    \includegraphics[scale=0.1]{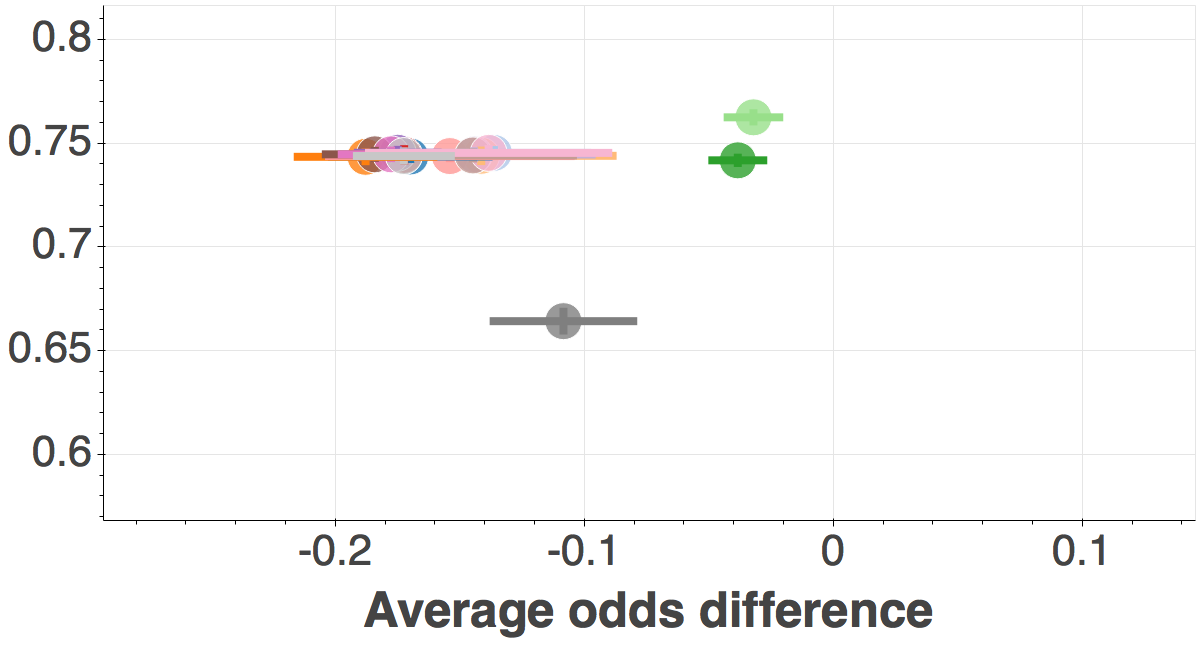}
    \includegraphics[scale=0.1]{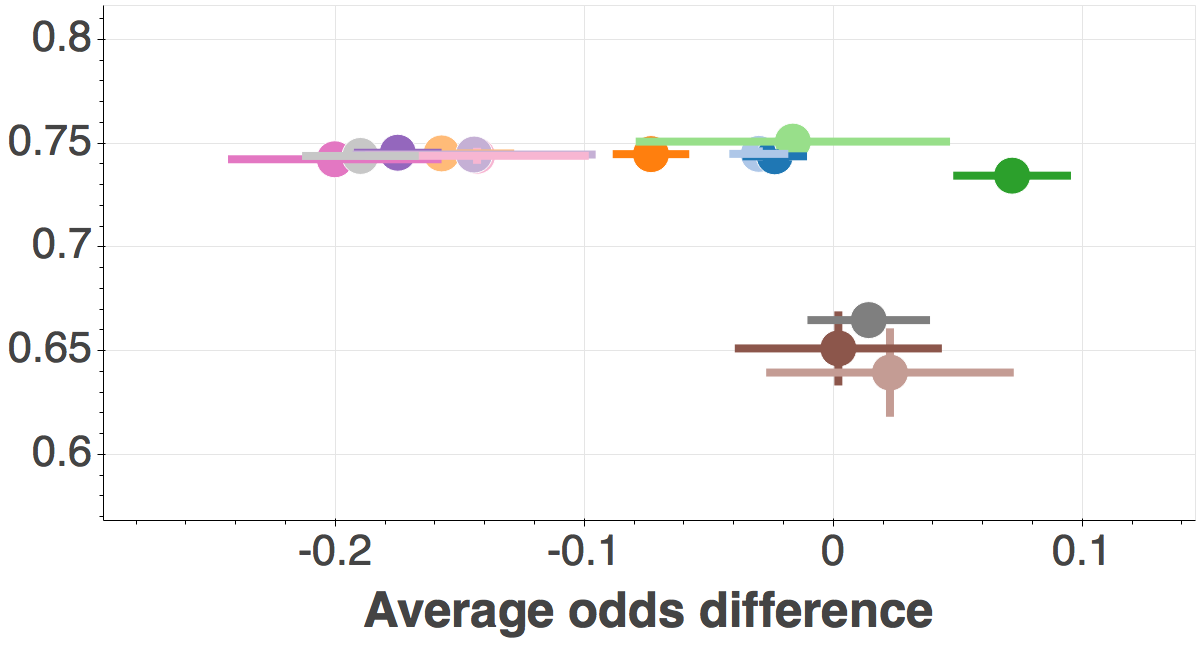}
    \label{fig:adult_race_average_odds_difference}
    \caption{Average odds difference}
\end{subfigure}
\begin{subfigure}{0.245\textwidth}
    \centering
    \includegraphics[scale=0.1]{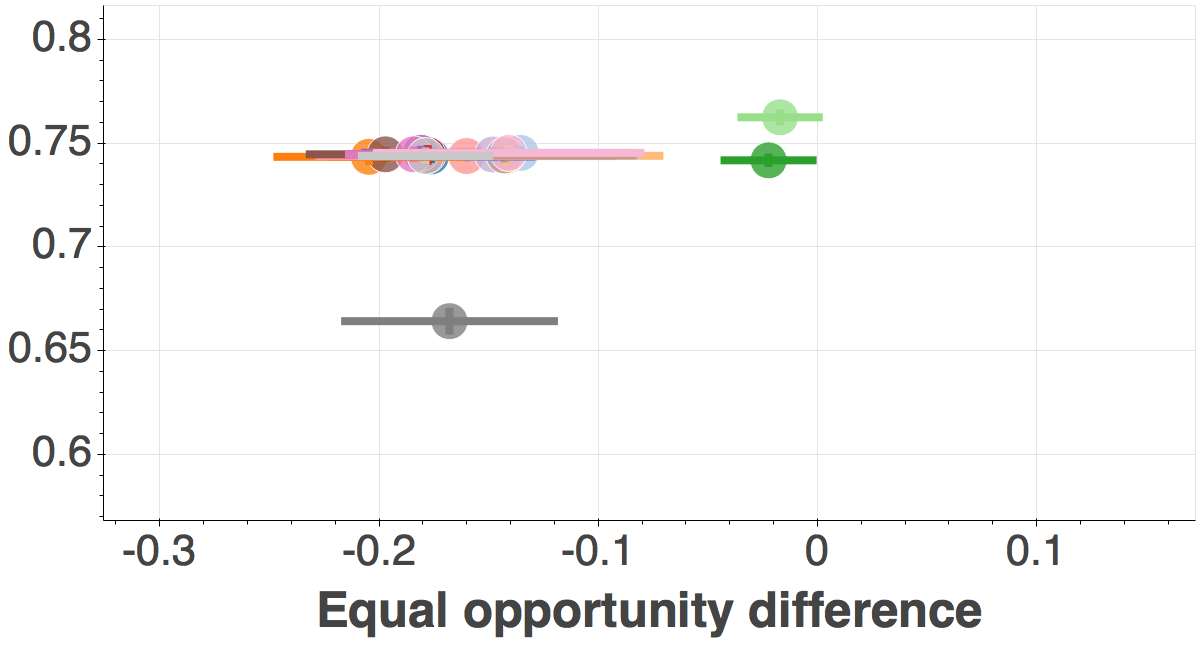}
    \includegraphics[scale=0.1]{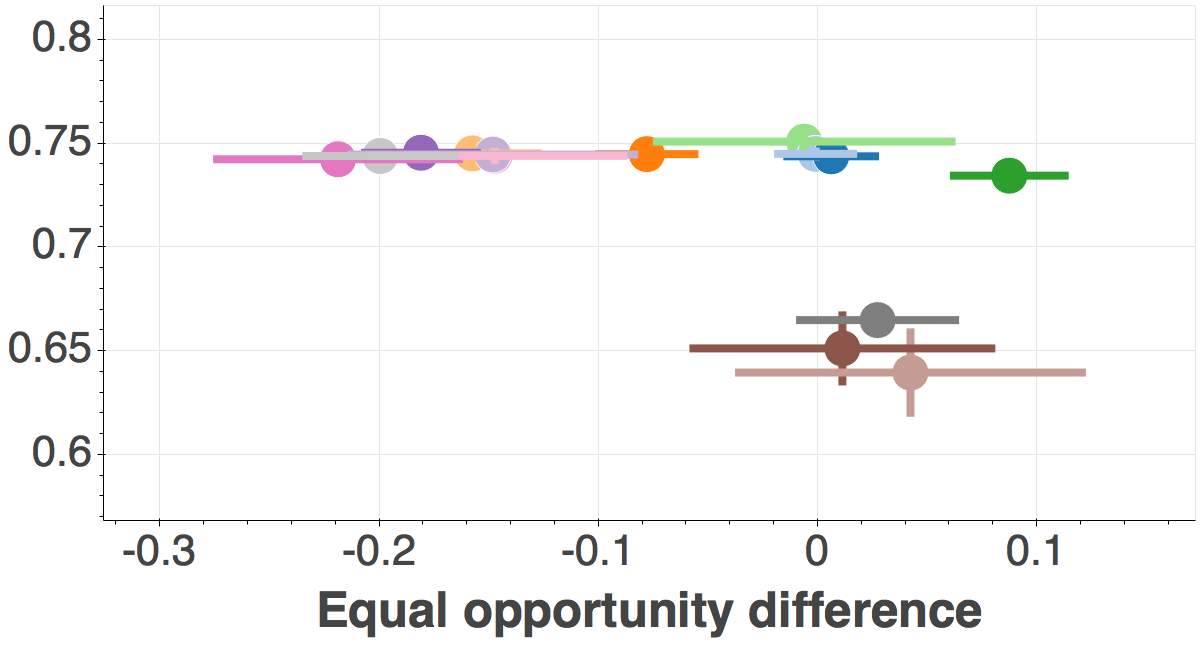}
    \label{fig:adult_race_equal_opportunity_difference}
    \caption{Equal opportunity difference}
\end{subfigure}
\caption{Fairness vs. Balanced Accuracy before (top panel) and after (bottom panel) applying various bias mitigation algorithms. Four different fairness metrics are shown. In most cases two classifiers (Logistic regression - LR or Random forest classifier - RF) were used. The ideal fair value of disparate impact is 1, whereas for all other metrics it is 0. The circles indicate the mean value and bars indicate the extent of $\pm$1 standard deviation. Dataset: \textit{Adult}, Protected attribute: \textit{race}.}
\label{fig:adult-race}
\end{scriptsize}
\end{figure*}
    






We also train classifiers with and without bias mitigation for all the dataset and algorithm combinations. We then measure statistical parity difference and disparate impact using the predicted dataset, and average odds and equal opportunity difference using the input and predicted datasets. In all possible cases, the validation datasets were used to obtain the score threshold for the classifiers to maximize balanced accuracy. The results are all obtained on test partitions.

An example result for \textit{Adult Census Income} dataset with \textit{race}
as protected attribute is shown in Figure \ref{fig:adult-race}. The rest of the results referenced here are available in Appendix \ref{appendix:evaluationgraphs}. Disparate impact remover and adversarial debiasing use differently processed datasets and hence their metrics in the top panel are different from others. The first thing that strikes when glancing at the figure is that the four different metrics seem to be correlated. Also the uncertainty in classification accuracy is much smaller compared to the uncertainty in the fairness metrics. There is also correlation between the metrics computed using the LR and RF classifiers. The Reject option classification algorithm improves fairness quite a bit, but also suffers a significant reduction in accuracy. Reweighing and optimized pre-processing also show good improvement in fairness, but without much penalty in accuracy. The two equal odds post-processing methods do not show significant changes to accuracy or fairness.

These evaluations can be used to arrive at an informed conclusion about the choice of bias mitigation algorithm depending on the application. Clearly, if modifying the data is possible reweighing or optimized pre-processing are good options.  However, if post-processing is the only option, e.g., if the user has access only to black box models, reject option classification is shown to be a decent choice on average, although it is brittle. The effect of dataset size on these metrics becomes clear looking at Figures \ref{fig:german-sex} and \ref{fig:german-age}. The uncertainty is substantially higher in the case of \textit{German Credit} data. Fairness improvements are more mixed in the \textit{COMPAS} data, but Reweighing and Reject option classification still show up to be good choices.


\section{Web Application}
\label{sec:webapp}

AIF360 includes not only the main toolkit code, but also an interactive Web experience (see Appendix \ref{appendix:posterpage} for a screen shot). Here we describe its front-end and back-end design.

\subsection{Design of the interactive experience}

The Web experience was designed to provide useful information for a diverse consumers. For business users, the interactive demonstration offers a sample of the toolkit's fairness checking and mitigation capabilities without requiring any programming knowledge. For new developers, the demonstration, notebook-based tutorials, guidance on algorithm selection, and access to a user community provide multiple ways to progress from their current level of understanding into the details of the code. For more advanced developers, the detailed documentation and code are directly accessible.

\begin{figure} [h!]
  \includegraphics[width=\linewidth]{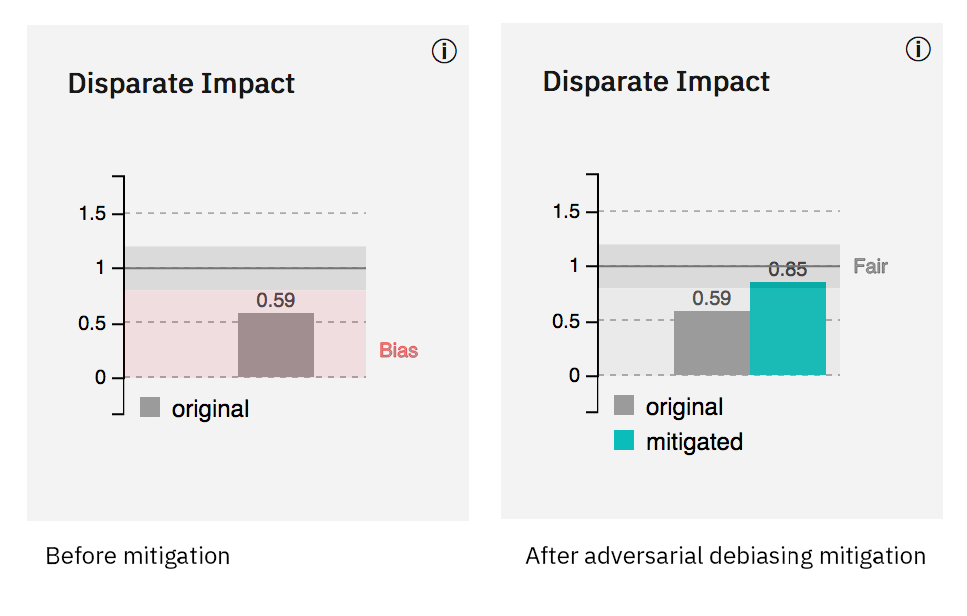}
  \caption{Graphs from the interactive web experience showing one of the metrics, for one of the datasets, before and after mitigation.}
  \label{fig:webui}
\end{figure}

The design of the Web experience proceeded through several iterations. Early clickable mock-ups of the interactive demonstration had users first select a dataset, one or two protected attributes to check for bias, and one of up to five metrics to use for checking. We learned, however, that this was overwhelming, even for those familiar with AI, since it required choices they were not yet equipped to make. As a result, we simplified the experience by asking users to first select only one of three datasets to explore. Bias checking results were then graphically presented for two protected attributes across five different metrics. Users could then select a mitigation algorithm leading to a report comparing bias before and after mitigation. The design of the charts for each bias metric also evolved in response to user feedback as we learned the importance of depicting a color-coded range of values considered fair or biased with more detailed information being available in an overlay. Figure \ref{fig:webui} shows the before and after mitigation graphs from the interactive Web experience. 

The design of the rest of the site also went through several iterations. Of particular concern, the front page sought to convey toolkit richness while still being approachable. In the final design, a short textual introduction to the content of the site, along with direct links to the API documentation and code repository, is followed by a number of direct links to various levels of advice and examples. Further links to the individual datasets, the bias checkers, and the mitigation algorithms are also provided. In all this, we ensured the site was suitably responsive across all major desktop and mobile platforms.

\subsection{Design of the back-end service}

The demo web application not only provides a gentle introduction to the capabilities of the toolkit, but also serves as an example of deploying the toolkit to the cloud and converting it into a web service. We used Python's Flask framework for building the service and exposed a REST API that generates a bias report based on the following input parameters from a user: the dataset name, the protected attributes, the privileged and unprivileged groups, the chosen fairness metrics, and the chosen mitigation algorithm, if any. With these inputs, the back-end then runs a series of steps to 1) split the dataset into training, development, and validation sets; 2) train a logistic regression classifier on the training set; 3) run the bias-checking metrics on the classifier against the test dataset; 4) if a mitigation algorithm is chosen, run the mitigation algorithm with the appropriate pipeline (pre-processing, in-processing, or post-processing). The end result is then cached so that if the exact same inputs are provided, the result can be directly retrieved from cache and no additional computation is needed.  

The reason to truly use the toolkit code in serving the Web application rather than having a pre-computed lookup table of results is twofold: we want to make the app a real representation of the underlying capabilities (in fact, creating the Web app helped us debug a few items in the code), and we also avoid any issues of synchronizing updates to the metrics, explainers, and algorithms with the results shown: synchronization is automatic. Currently, the service is limited to three built-in datasets, but it can be expanded to support the user's own data upload. The service is also limited to building logistic regression classifiers, but again this can be expanded. Such expansions can be more easily implemented if this fairness service is integrated into a full AI suite that provides various classifier options and data storage solutions.

\section{Conclusion}
\label{sec:discussion}

AIF360 is an open source toolkit that brings value to diverse users and practitioners.  For fairness researchers, it provides a platform that enables them to: 1) experiment with and compare various existing bias detection and mitigation algorithms in a common framework, and gain insights into their practical usage; 2) contribute and benchmark new algorithms; 3) contribute new datasets and analyze them for bias.  For developers, it provides: 1) education on the important issues in bias checking and mitigation, 2) guidance on which metrics and mitigation algorithms to use; 3) tutorials and sample notebooks that demonstrate bias mitigation in different industry settings; and 4) a Python package for detecting and mitigating bias in their workflows.

Fairness is a multifaceted, context-dependent social construct that defies simple definition. The metrics and algorithms in AIF360 may be viewed from the lens of distributive justice \citep{HuC2018}, i.e., relating to decisions about who in a society receives which benefits and goods, and clearly do not capture the full scope of fairness in all situations. Even within distributive justice, more work is needed to apply the toolkit to additional datasets and situations. Future work could also expand the toolkit to measure and mitigate other aspects of justice such as compensatory justice, i.e. relating to the extent to which people are fairly compensated for harms done to them. Further work is also needed to extend the variety of types of explanations offered, and to create guidance for practitioners on when a specific kind of explanation is most appropriate. There is a lot of work left to do to achieve unbiased AI, we hope others in the research community continue to contribute to the toolkit their own approaches to fairness and bias checking, mitigation and explanation. 

\bibliographystyle{sysml2019} 
\bibliography{sysml2019}

\onecolumn
\appendixpage
\appendix
\section{UML Class Diagram}
\label{appendix:uml}
\begin{figure*}[h!]
    \centering
    \includegraphics[width=\textwidth]{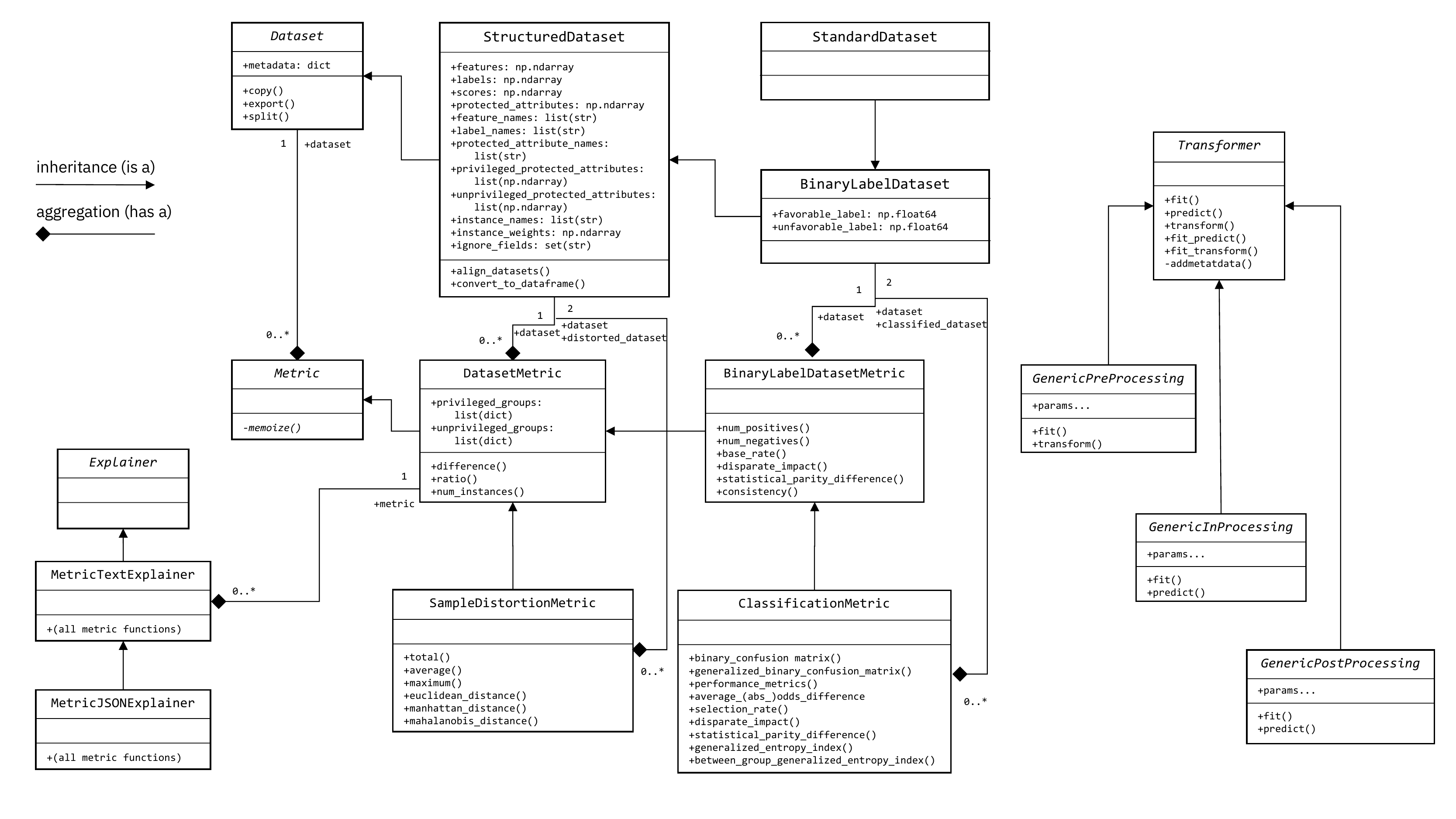}
    \caption{Class abstractions for a fair machine learning pipeline, as implemented in AIF360. This figure is meant to provide a visual sense of the class hierarchy, many details and some methods are omitted. For brevity, inherited members and methods are not shown (but overridden ones are) nor are aliases such as \texttt{recall()} for \texttt{true\_positive\_rate()}. Some methods are ``meta-metrics'' --- such as \texttt{difference()}, \texttt{ratio()}, \texttt{total()}, \texttt{average()}, \texttt{maximum()} --- that act on other metrics to get, e.g. \texttt{true\_positive\_rate\_difference()}. The metric explainer classes use the same method signatures as the metric classes (not enumerated) but provide further description for the values. The \texttt{GenericPreProcessing}, \texttt{GenericInProcessing}, and \texttt{GenericPostProcessing} are not actual classes but serve as placeholders here for the real bias mitigation algorithms we implemented. Finally, \texttt{memoize} and \texttt{addmetadata} are Python decorator functions that are automatically applied to every function in their respective classes.}
    \label{fig:uml_diagram}
\end{figure*}

\section{Code Snippets}
\label{appendix:snippets}
This example provides Python code snippets for some common tasks that the user might perform using our toolbox. The example involves the user loading a dataset, splitting it into training and testing partitions, understanding the outcome disparity between two demographic groups, and transforming the dataset to mitigate this disparity. A more detailed version of this example is available in \url{url.redacted}. 


\subsection{Dataset operations}
\begin{verbatim}
# Load the UCI Adult dataset
from aif360.datasets import AdultDataset
ds_orig = AdultDataset()

# Split into train and test partitions
ds_orig_tr, ds_orig_te = ds_orig.split([0.7], shuffle=True, seed=1)

# Look into the training dataset
print("Training Dataset shape")
print(ds_orig_tr.features.shape)
print("Favorable and unfavorable outcome labels")
print(ds_orig_tr.favorable_label, ds_orig_tr.unfavorable_label)
print("Metadata for labels")
print(ds_orig_tr.metadata["label_maps"])
print("Protected attribute names")
print(ds_orig_tr.protected_attribute_names)
print("Privileged and unprivileged protected attribute values")
print(ds_orig_tr.privileged_protected_attributes, 
    ds_orig_tr.unprivileged_protected_attributes)
print("Metadata for protected attributes")
print(ds_orig_tr.metadata["protected_attribute_maps"])

\end{verbatim}

\paragraph{Expected output:} The attributes of the \texttt{Adult} dataset will be printed. The training partition of the \textit{Adult} dataset has 31655 instances and 98 features with two protected attributes (\texttt{race} and \texttt{sex}). The labels correspond to high-income ($>50$K) or low-income ($<=50$K), as shown in the metadata. Similar metadata is also available for protected attributes.

\subsection{Checking for bias in the original data}
\begin{verbatim}
# Load the metric class
from aif360.metrics import BinaryLabelDatasetMetric

# Define privileged and unprivileged groups
priv = [{'sex': 1}] # Male
unpriv = [{'sex': 0}] # Female

# Create the metric object
metric_otr = BinaryLabelDatasetMetric( ds_orig_tr, 
    unprivileged_groups=unpriv, privileged_groups=priv)
        
# Load and create explainers
from aif360.explainers import MetricTextExplainer, MetricJSONExplainer
text_exp_otr = MetricTextExplainer(metric_otr)
json_exp_otr = MetricJSONExplainer(metric_otr)
        
# Print statistical parity difference
print(text_exp_otr.statistical_parity_difference())
print(json_exp_otr.statistical_parity_difference())
\end{verbatim}

\paragraph{Expected output:} The statistical parity difference should be $-0.1974$, which is the difference between probability of favorable outcome (high income) between the unprivileged group (females) and the privileged group (male) in this dataset. The JSON output is more elaborate to facilitate consumption by a downstream algorithm.

\subsection{Pre-process data to mitigate bias}
\begin{verbatim}
# Import the reweighing preprocessing algorithm class
from aif360.algorithms.preprocessing.reweighing import Reweighing
    
# Create the algorithm object
RW = Reweighing(unprivileged_groups=unpriv, privileged_groups=priv)
    
# Train and predict on the training data
# Uses scikit-learn convention (fit, predict, transform)
RW.fit(ds_orig_tr)
ds_transf_tr = RW.transform(ds_orig_tr)
\end{verbatim}

\paragraph{Expected output:} There will be no output here, but the reweighing algorithm equalizes the weights across (group, label) combination. 

\subsection{Checking for bias in the pre-processed training data}
\begin{verbatim}
# Create the metric object for pre-processed data
metric_ttr = BinaryLabelDatasetMetric(ds_transf_tr,
    unprivileged_groups=unpriv, privileged_groups=priv)
        
# Create explainer
text_exp_ttr = MetricTextExplainer(metric_ttr)

# Print statistical parity difference
print(text_exp_ttr.statistical_parity_difference())
\end{verbatim}

\paragraph{Expected output:} Because of the action of the re-weighing pre-processing algorithm, the statistical parity difference for the transformed data (\texttt{ds\_transf\_tr}) must be really close to 0.

\subsection{Pre-process out-of-sample testing data and check for bias}
\begin{verbatim}
# Apply the learned re-weighing pre-processor
ds_transf_te = RW.transform(ds_orig_te)

# Create metric objects for original and 
# pre-processed test data
metric_ote = BinaryLabelDatasetMetric(ds_orig_te,
    unprivileged_groups=unpriv, privileged_groups=priv)
metric_tte = BinaryLabelDatasetMetric(ds_transf_te,
    unprivileged_groups=unpriv, privileged_groups=priv)

# Create explainers for both metric objects
text_exp_ote = MetricTextExplainer(metric_ote)
text_exp_tte = MetricTextExplainer(metric_tte)

# Print statistical parity difference
print(text_exp_ote.statistical_parity_difference())
print(text_exp_tte.statistical_parity_difference())
\end{verbatim}

\paragraph{Expected output:} The trained re-weighing pre-processor can be applied on the out-of-sample test data. The metrics for the original and transformed testing data will show a significant reduction in statistical parity difference (-0.2021 to -0.0119 in this case).

\newpage

\section{Additional Experimental Details}
\label{appendix:expt_details}
We provide additional details on the experimental evaluations.

\subsection{Datasets}
\subsubsection{Adult Census Income}
For protected attribute \textit{sex}, \textit{Male} is privileged, and \textit{Female} is unprivileged. For protected attribute \textit{race}, \textit{White} is privileged, and \textit{Non-white} is unprivileged. Favorable label is \textit{High income} ($>50$K) and unfavorable label is \textit{Low income} ($<=50$K).

\subsubsection{German Credit}
For protected attribute \textit{sex}, \textit{Male} is privileged, and \textit{Female} is unprivileged. For protected attribute \textit{age}, \textit{Old} is privileged, and \textit{Young} is unprivileged. Favorable label is \textit{Good credit} and unfavorable label is \textit{Bad credit}.
 
\subsubsection{Probpublica recidivism (COMPAS)}
For protected attribute \textit{sex}, \textit{Female} is privileged, and \textit{Male} is unprivileged. For protected attribute \textit{race}, \textit{Caucasian} is privileged, and \textit{Not Caucasian} is unprivileged. Favorable label is \textit{Did not recidivate} and unfavorable label is \textit{Did recidivate}.

\subsection{Metrics}
\subsubsection{Statistical Parity Difference}
This is the difference in the probability of favorable outcomes between the unprivileged and privileged groups. This can be computed both from the input dataset as well as from the dataset output from a classifier (predicted dataset). A value of $0$ implies both groups have equal benefit, a value less than $0$ implies higher benefit for the privileged group, and a value greater than $0$ implies higher benefit for the unprivileged group.

\subsubsection{Disparate Impact}
This is the ratio in the probability of favorable outcomes between the unprivileged and privileged groups. This can be computed both from the input dataset as well as from the dataset output from a classifier (predicted dataset). A value of $1$ implies both groups have equal benefit, a value less than $1$ implies higher benefit for the privileged group, and a value greater than $1$ implies higher benefit for the unprivileged group.

\subsubsection{Average odds difference}
This is the average of difference in false positive rates and true positive rates between unprivileged and privileged groups. This is a method in the \texttt{ClassificationMetric} class and hence needs to be computed using the input and output datasets to a classifier. A value of $0$ implies both groups have equal benefit, a value less than $0$ implies higher benefit for the privileged group and a value greater than $0$ implies higher benefit for the unprivileged group. 

\subsubsection{Equal opportunity difference}
This is the difference in true positive rates between unprivileged and privileged groups. This is a method in the \texttt{ClassificationMetric} class and hence needs to be computed using the input and output datasets to a classifier. A value of $0$ implies both groups have equal benefit, a value less than $0$ implies higher benefit for the privileged group and a value greater than $0$ implies higher benefit for the unprivileged group.

\section{Evaluation on different data sets}
\label{appendix:evaluationgraphs}
We present additional results with bias mitigation obtained for various datasets and protected attributes. These correspond to the setting described in Section \ref{sec:eval_algs}.


\begin{figure*}[h!]
\begin{scriptsize}
\begin{subfigure}{1.0\textwidth}
    \centering
    \includegraphics[scale=0.25]{legend_hor.png}
\end{subfigure}
\begin{subfigure}{0.245\textwidth}
    \centering
    \includegraphics[scale=0.1]{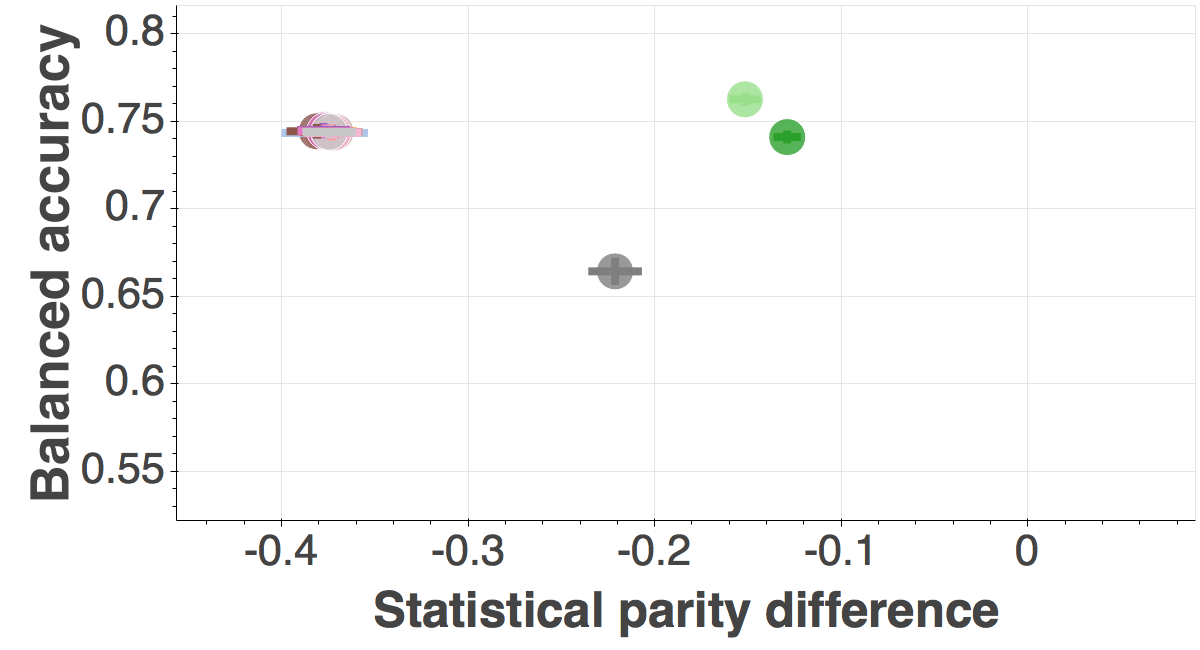}
    \includegraphics[scale=0.1]{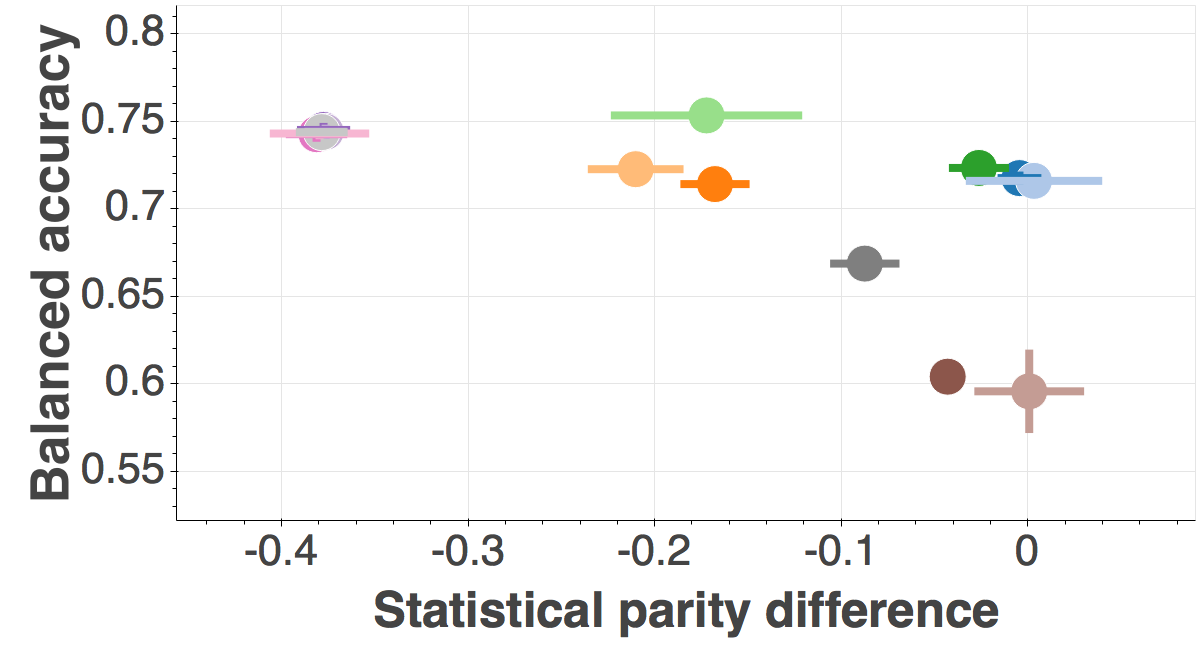}
    \label{fig:adult_sex_statistical_parity_difference}
    \caption{Statistical parity difference}
\end{subfigure}
\begin{subfigure}{0.245\textwidth}
    \centering
    \includegraphics[scale=0.1]{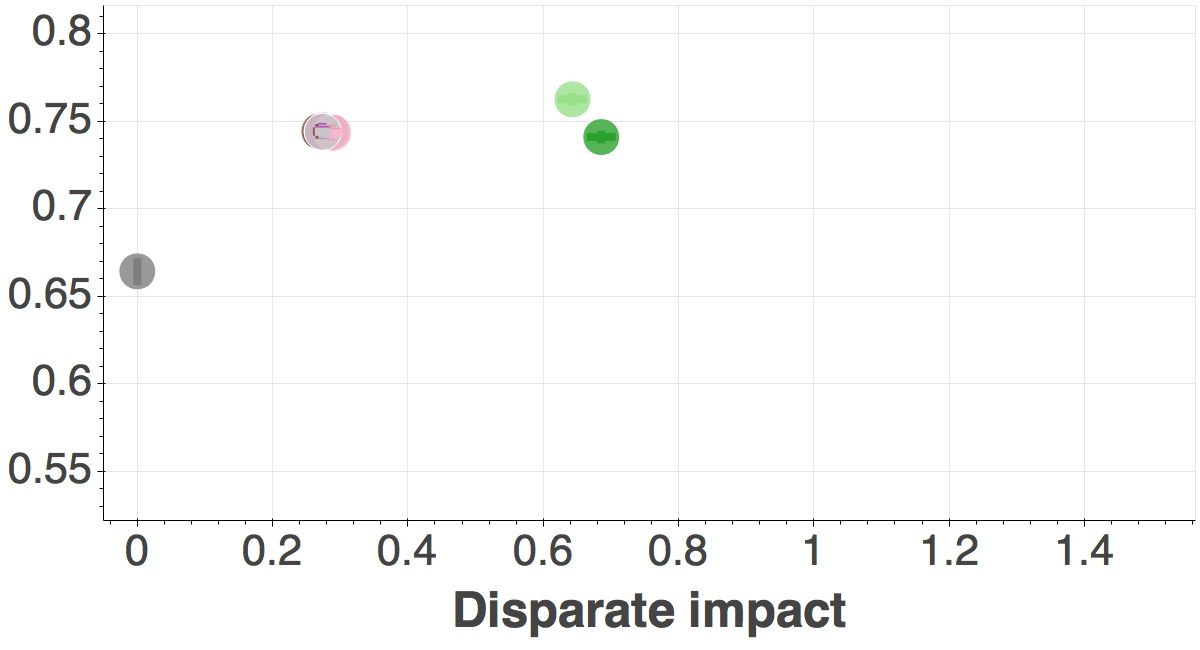}
    \includegraphics[scale=0.1]{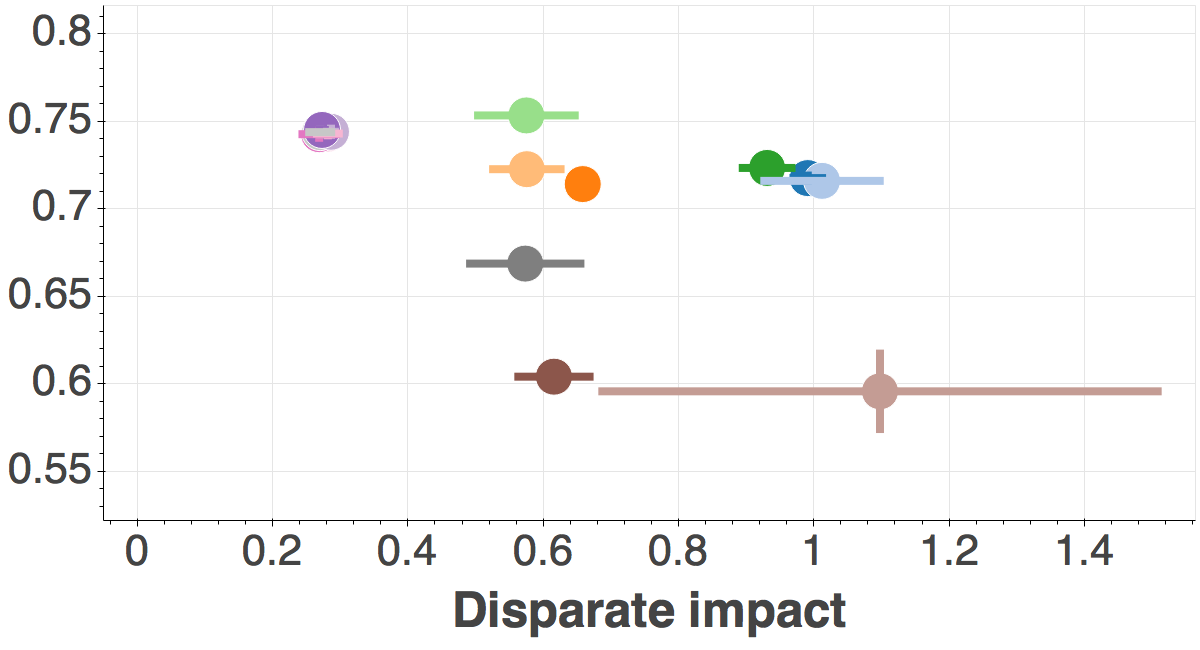}
    \label{fig:adult_sex_disparate_impact}
    \caption{Disparate impact}
\end{subfigure}
\begin{subfigure}{0.245\textwidth}
    \centering
    \includegraphics[scale=0.1]{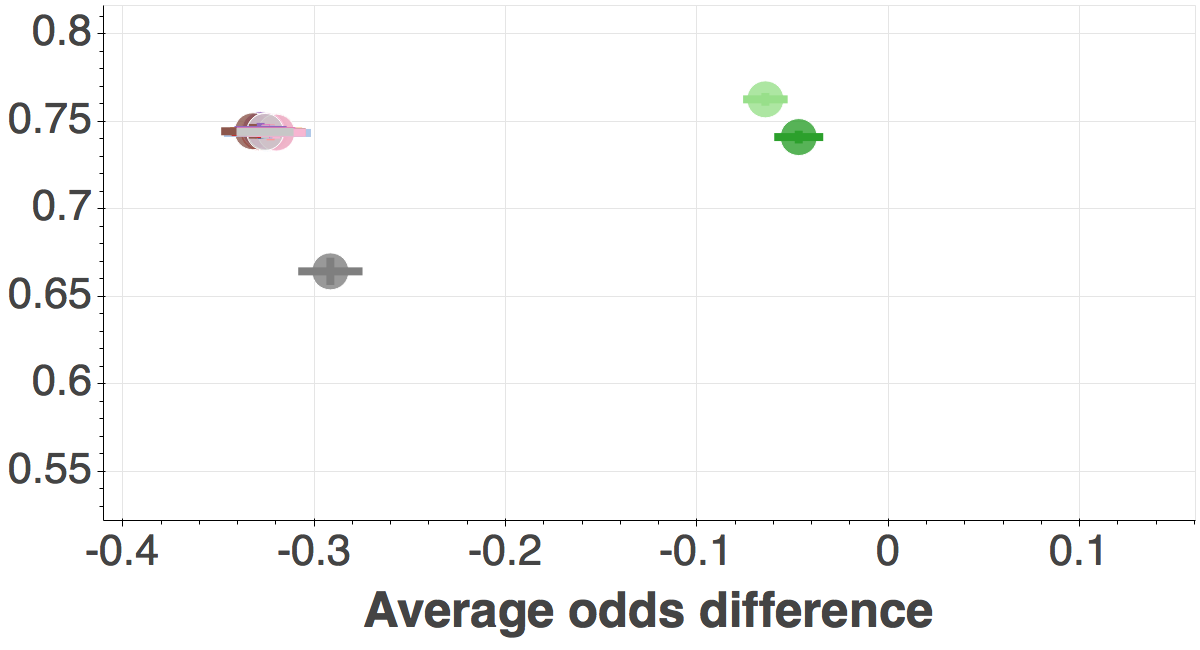}
    \includegraphics[scale=0.1]{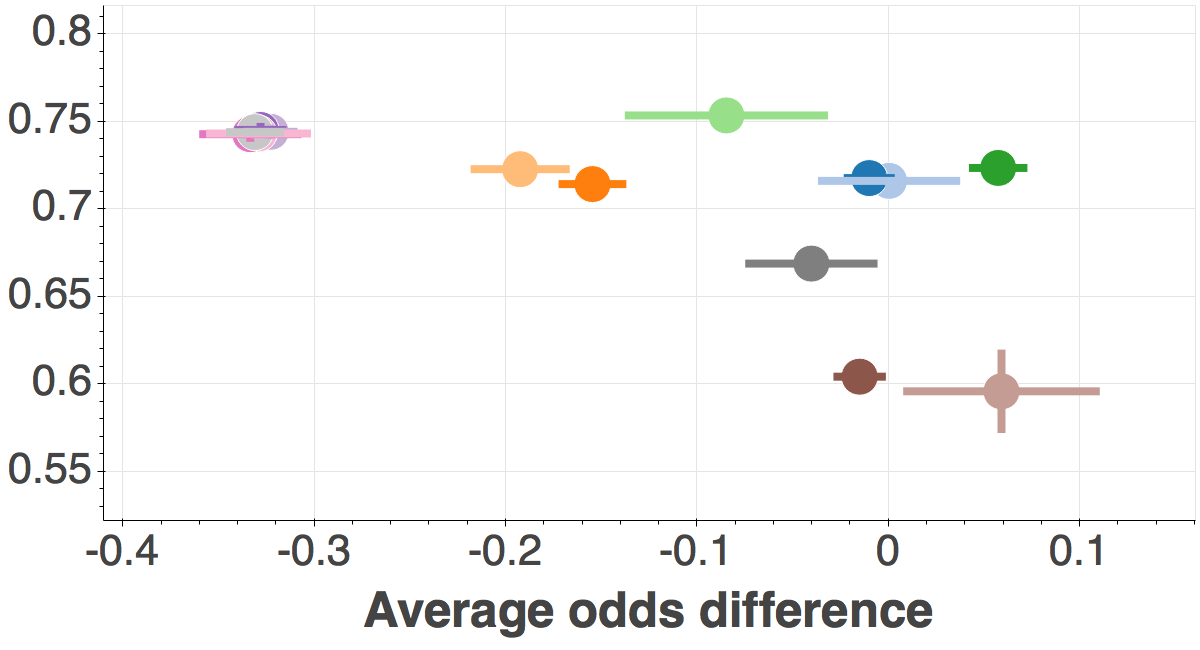}
    \label{fig:adult_sex_average_odds_difference}
    \caption{Average odds difference}
\end{subfigure}
\begin{subfigure}{0.245\textwidth}
    \centering
    \includegraphics[scale=0.1]{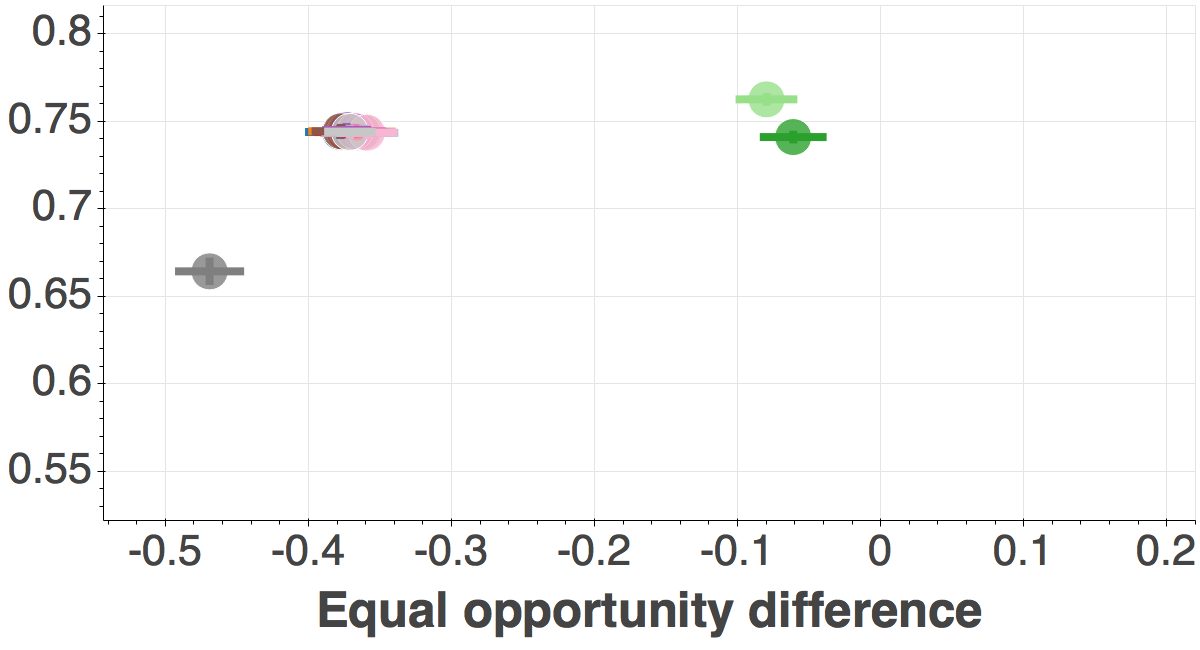}
    \includegraphics[scale=0.1]{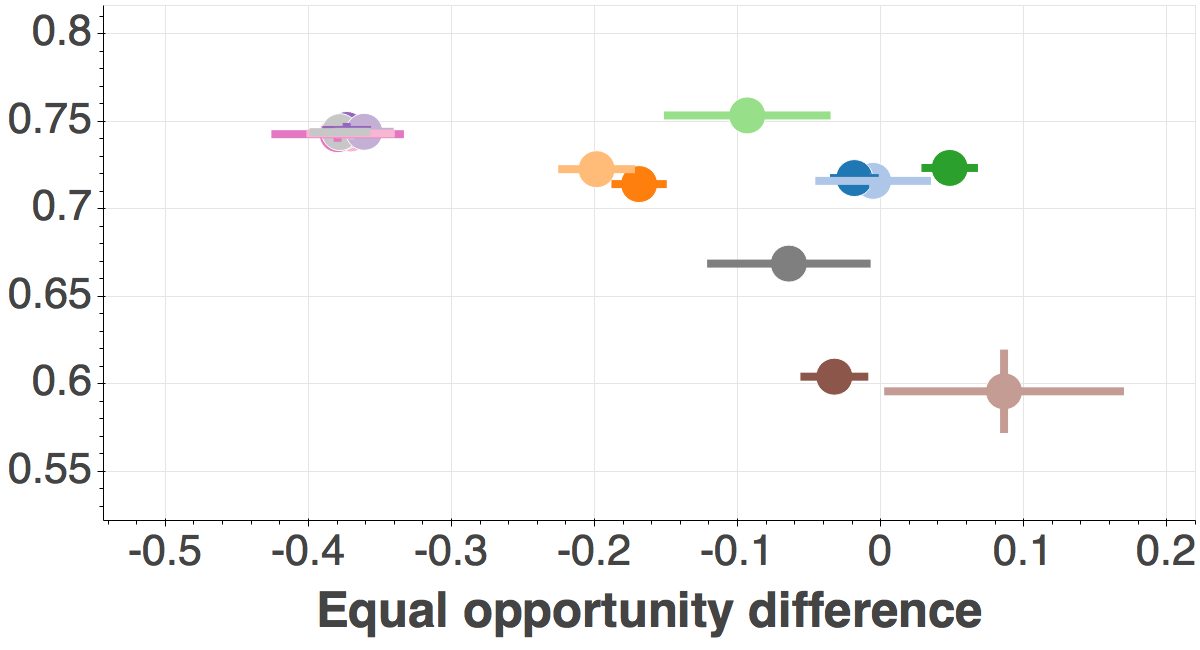}
    \label{fig:adult_sex_equal_opportunity_difference}
    \caption{Equal opportunity difference}
\end{subfigure}
\caption{Fairness vs. Balanced Accuracy before (top panel) and after (bottom panel) applying various bias mitigation algorithms. Four different fairness metrics are shown. In most cases two classifiers (Logistic regression - LR or Random forest classifier - RF) were used. The ideal fair value of disparate impact is 1, whereas for all other metrics it is 0. The circles indicate the mean value and bars indicate the extent of $\pm$1 standard deviation. Data set: \textit{Adult}, Protected attribute: \textit{sex}. }
\label{fig:adult-sex}
\end{scriptsize}
\end{figure*}


\begin{figure*}[h!]
\begin{scriptsize}
\begin{subfigure}{1.0\textwidth}
    \centering
    \includegraphics[scale=0.25]{legend_hor.png}
\end{subfigure}
\begin{subfigure}{0.245\textwidth}
    \centering
    \includegraphics[scale=0.1]{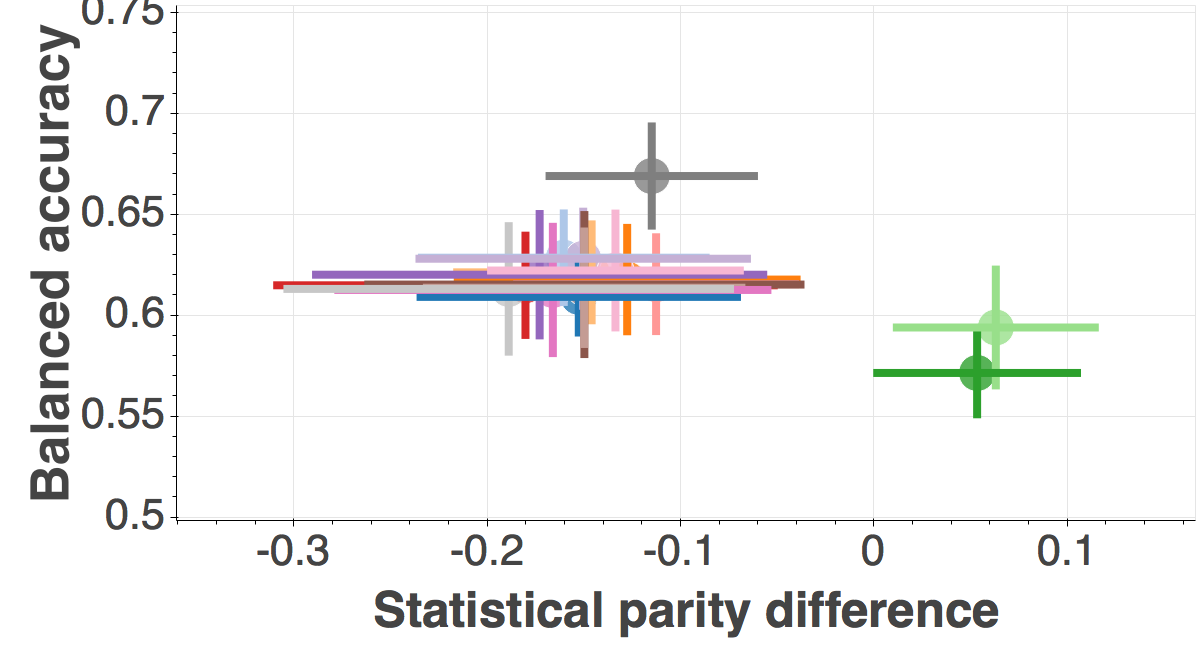}
    \includegraphics[scale=0.1]{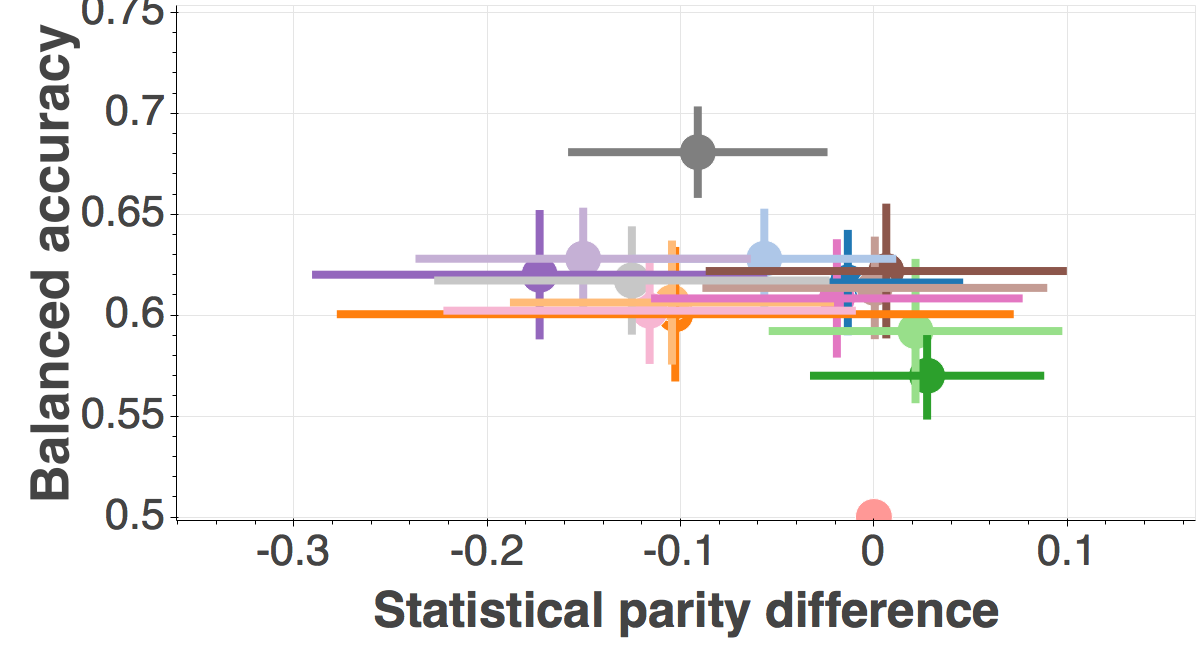}
    \label{fig:german_sex_statistical_parity_difference}
    \caption{Statistical parity difference}
\end{subfigure}
\begin{subfigure}{0.245\textwidth}
    \centering
    \includegraphics[scale=0.1]{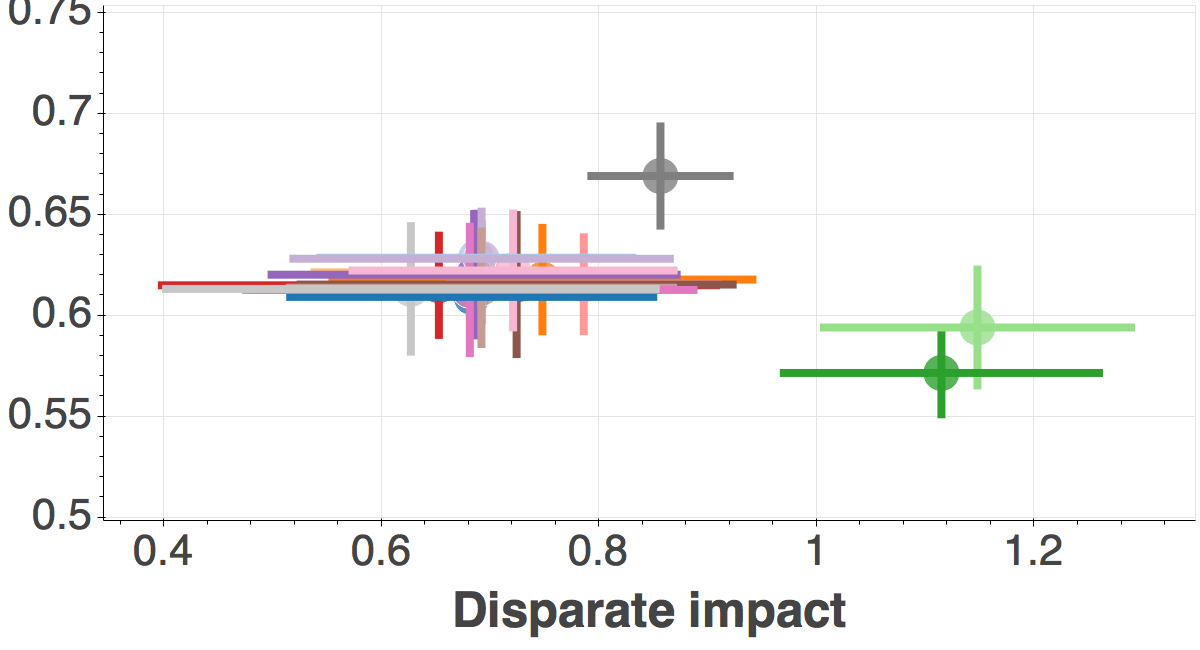}
    \includegraphics[scale=0.1]{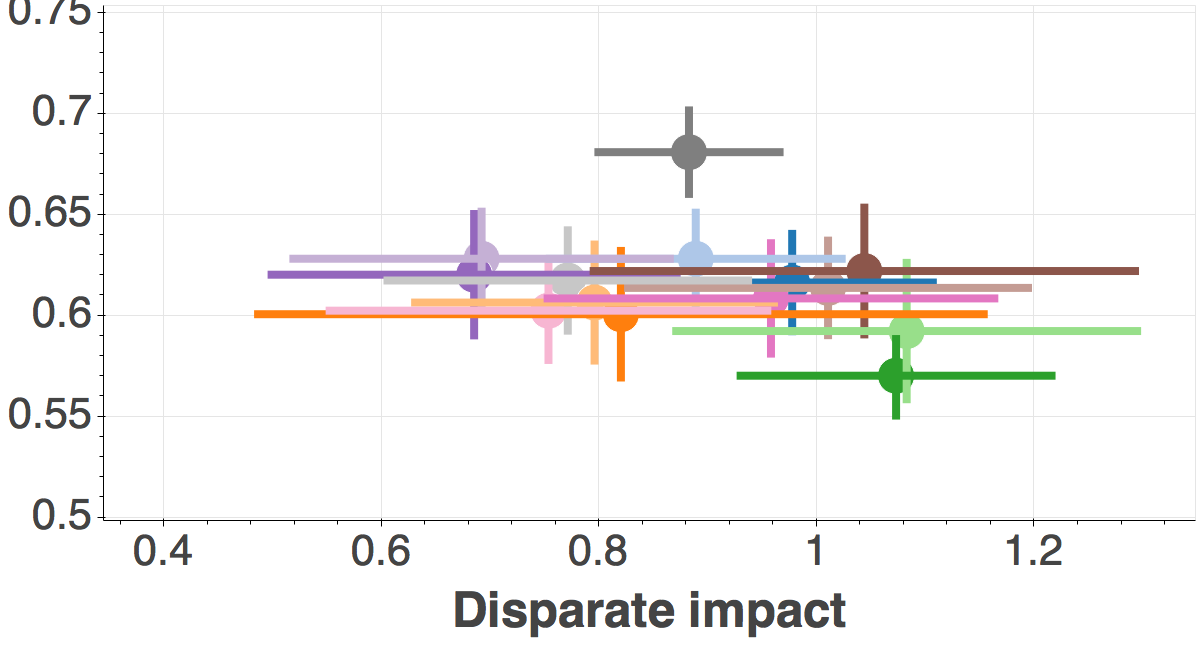}
    \label{fig:german_sex_disparate_impact}
    \caption{Disparate impact}
\end{subfigure}
\begin{subfigure}{0.245\textwidth}
    \centering
    \includegraphics[scale=0.1]{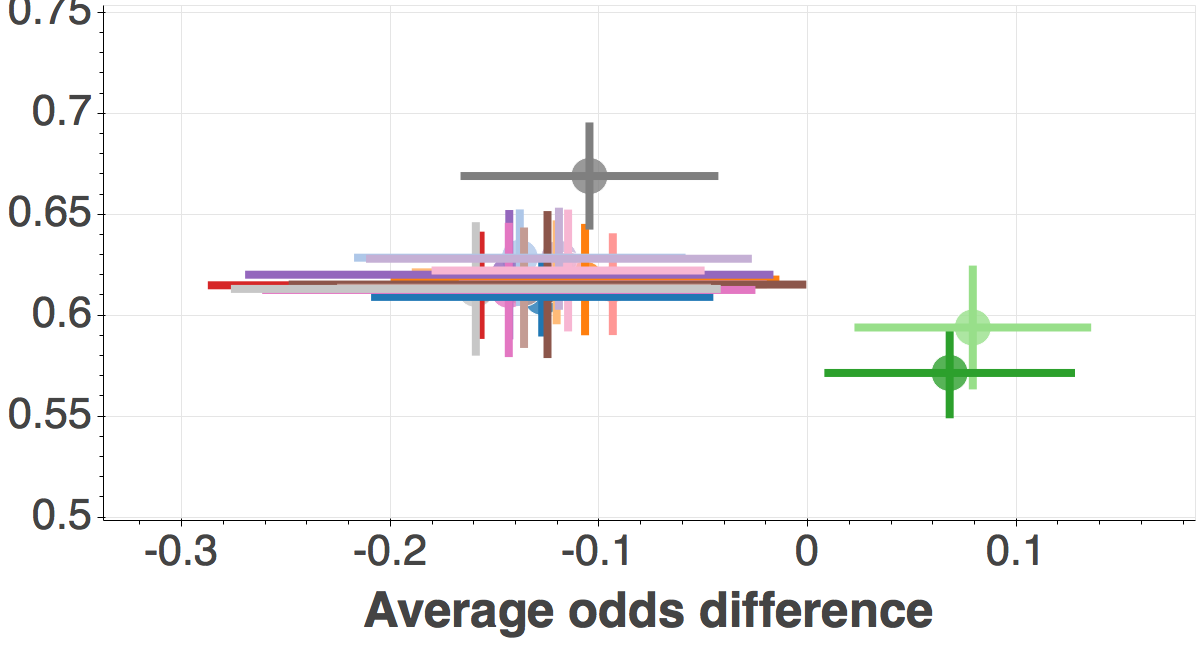}
    \includegraphics[scale=0.1]{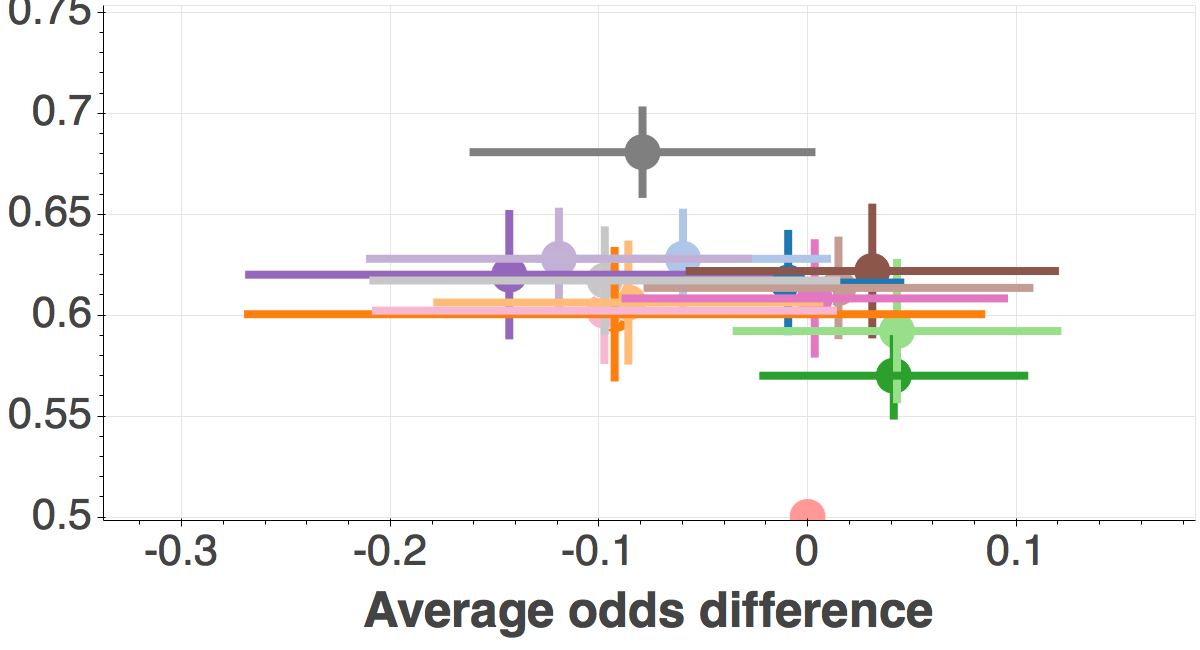}
    \label{fig:german_sex_average_odds_difference}
    \caption{Average odds difference}
\end{subfigure}
\begin{subfigure}{0.245\textwidth}
    \centering
    \includegraphics[scale=0.1]{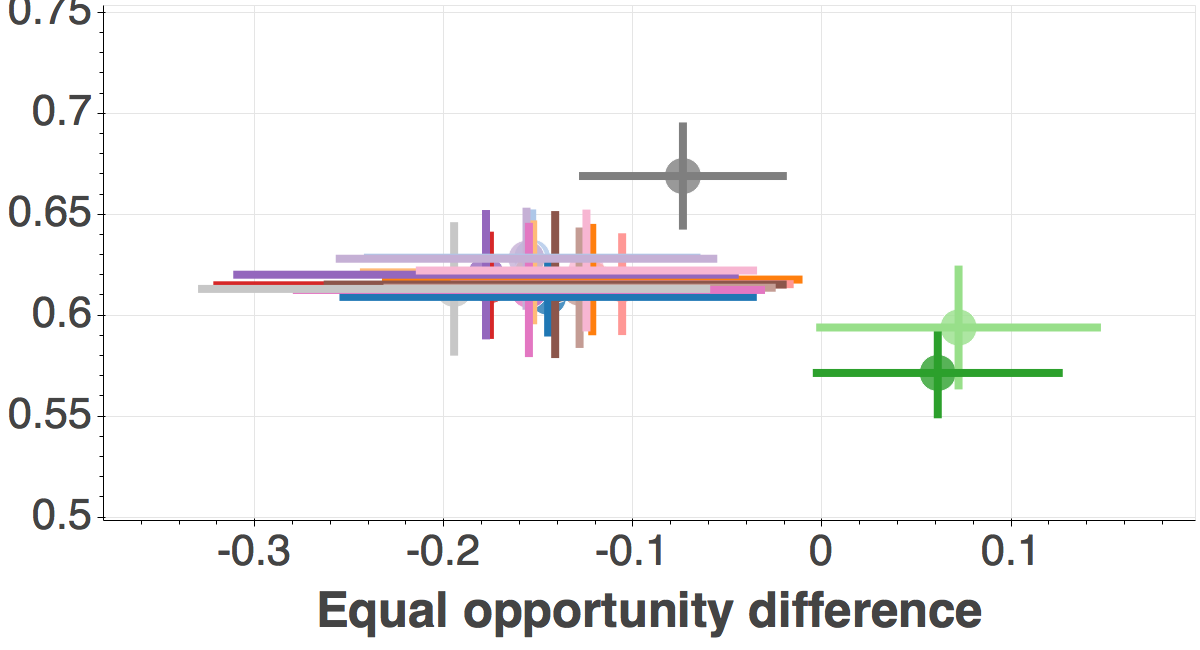}
    \includegraphics[scale=0.1]{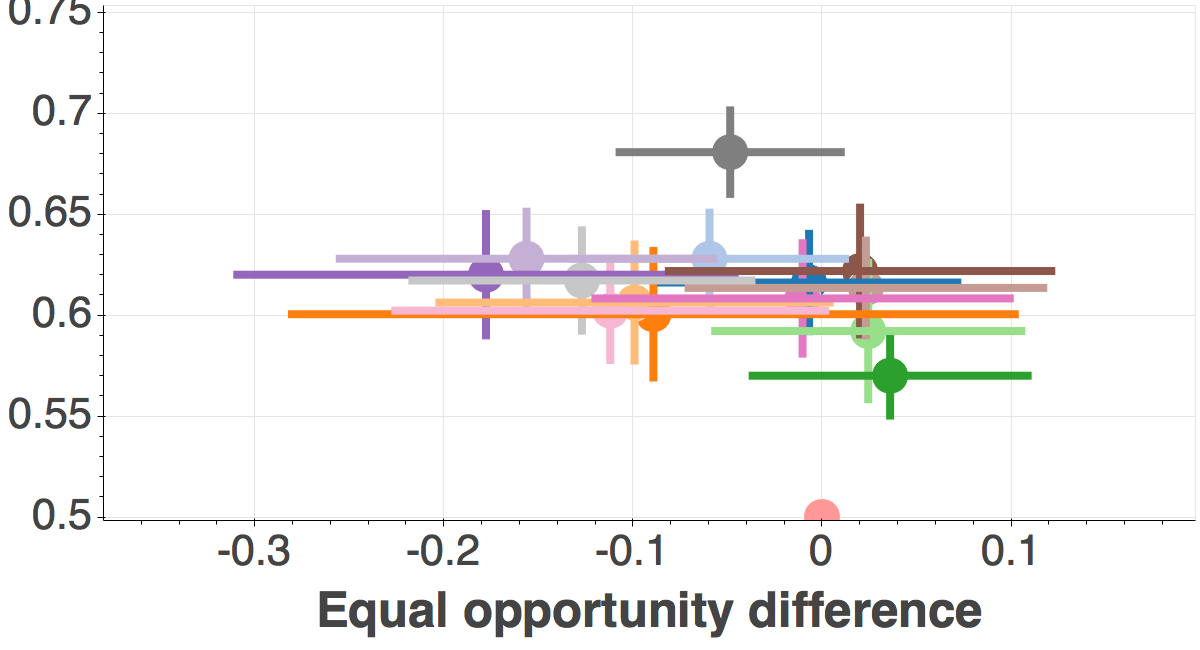}
    \label{fig:german_sex_equal_opportunity_difference}
    \caption{Equal opportunity difference}
\end{subfigure}
\caption{Fairness vs. Balanced Accuracy before (top panel) and after (bottom panel) applying various bias mitigation algorithms. Four different fairness metrics are shown. In most cases two classifiers (Logistic regression - LR or Random forest classifier - RF) were used. The ideal fair value of disparate impact is 1, whereas for all other metrics it is 0. The circles indicate the mean value and bars indicate the extent of $\pm$1 standard deviation. Data set: \textit{german}, Protected attribute: \textit{sex}. }
\label{fig:german-sex}
\end{scriptsize}
\end{figure*}


\begin{figure*}[h!]
\begin{scriptsize}
\begin{subfigure}{1.0\textwidth}
    \centering
    \includegraphics[scale=0.245]{legend_hor.png}
\end{subfigure}
\begin{subfigure}{0.24\textwidth}
    \centering
    \includegraphics[scale=0.1]{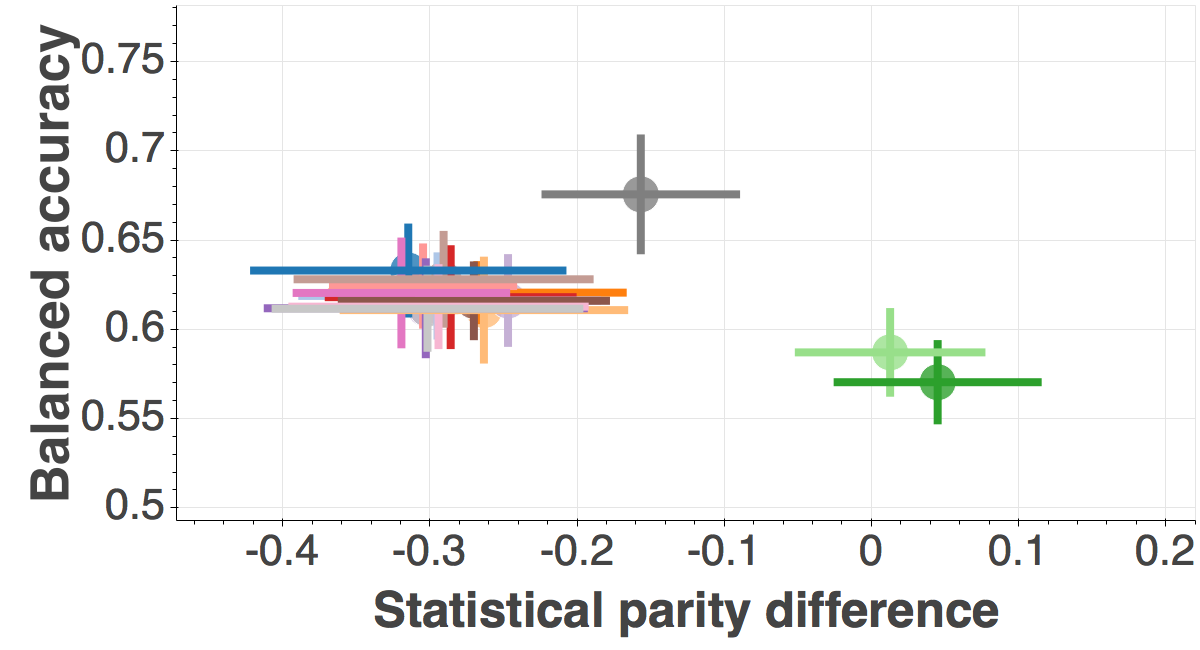}
    \includegraphics[scale=0.1]{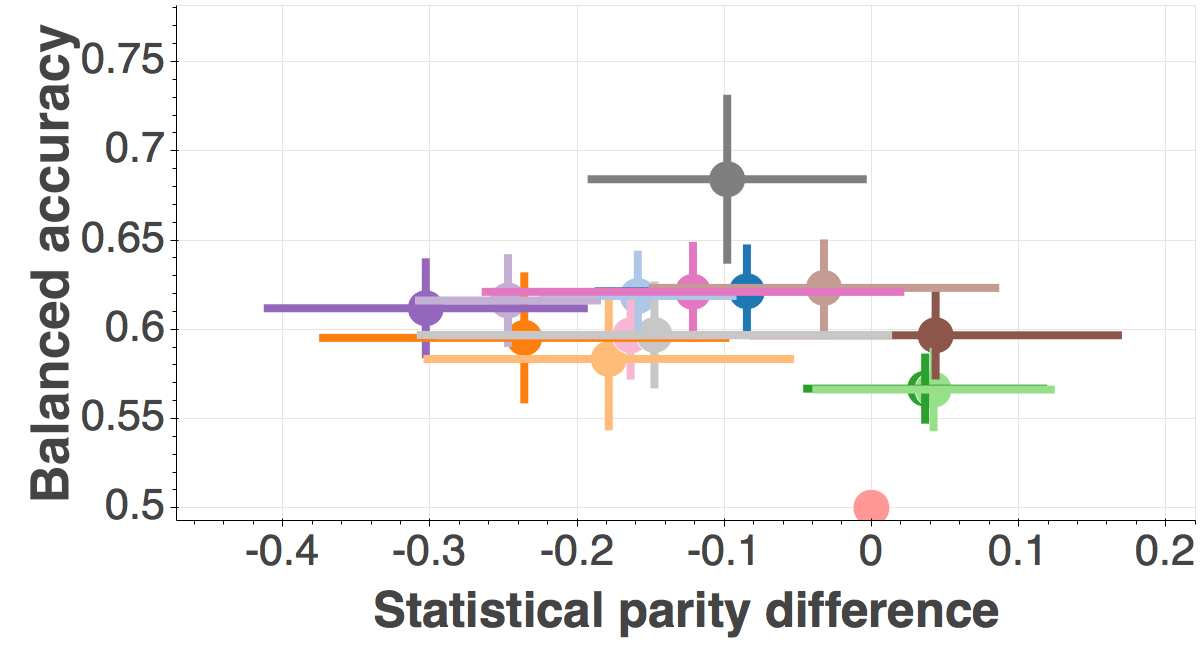}
    \label{fig:german_age_statistical_parity_difference}
    \caption{Statistical parity difference}
\end{subfigure}
\begin{subfigure}{0.245\textwidth}
    \centering
    \includegraphics[scale=0.1]{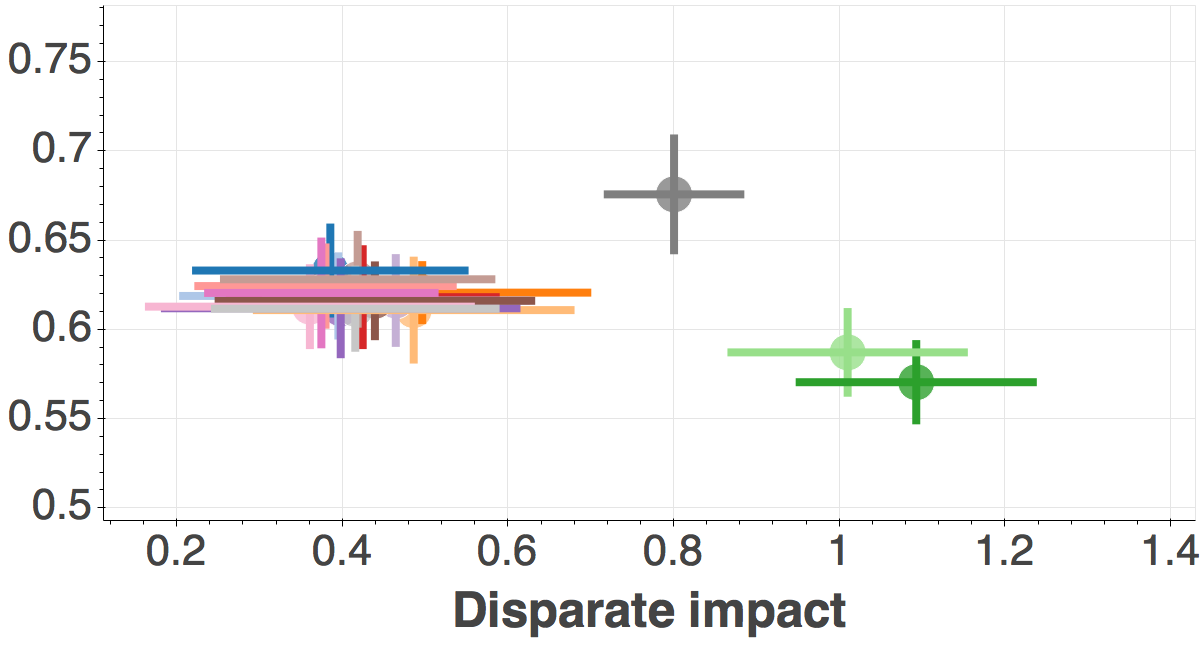}
    \includegraphics[scale=0.1]{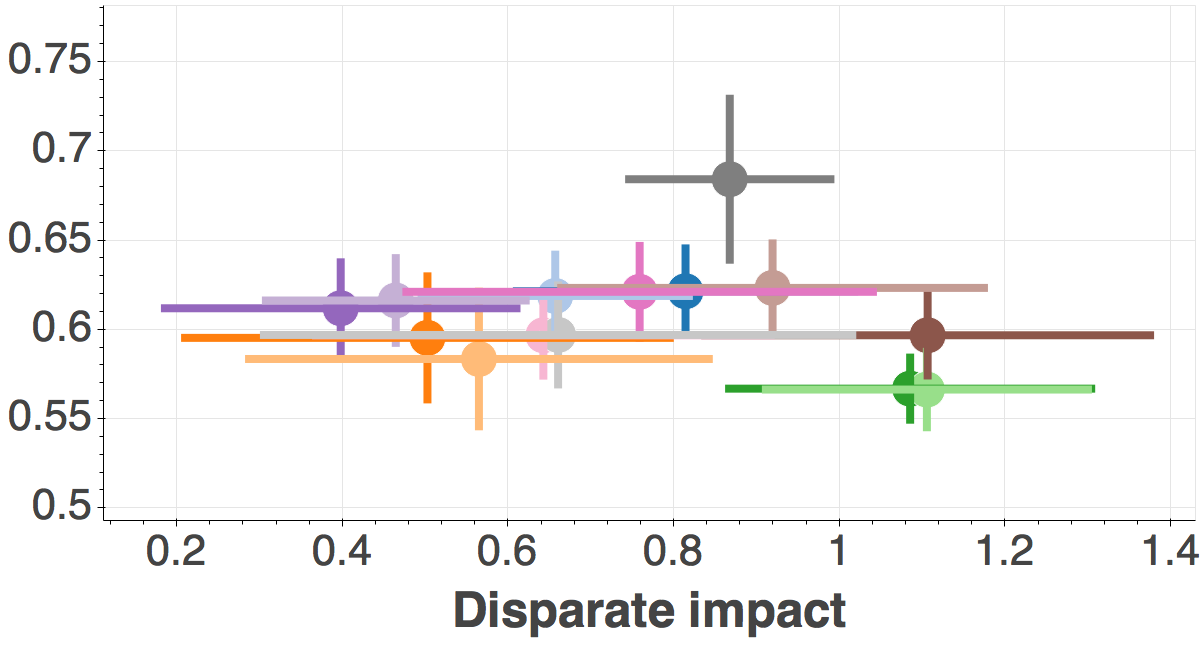}
    \label{fig:german_age_disparate_impact}
    \caption{Disparate impact}
\end{subfigure}
\begin{subfigure}{0.245\textwidth}
    \centering
    \includegraphics[scale=0.1]{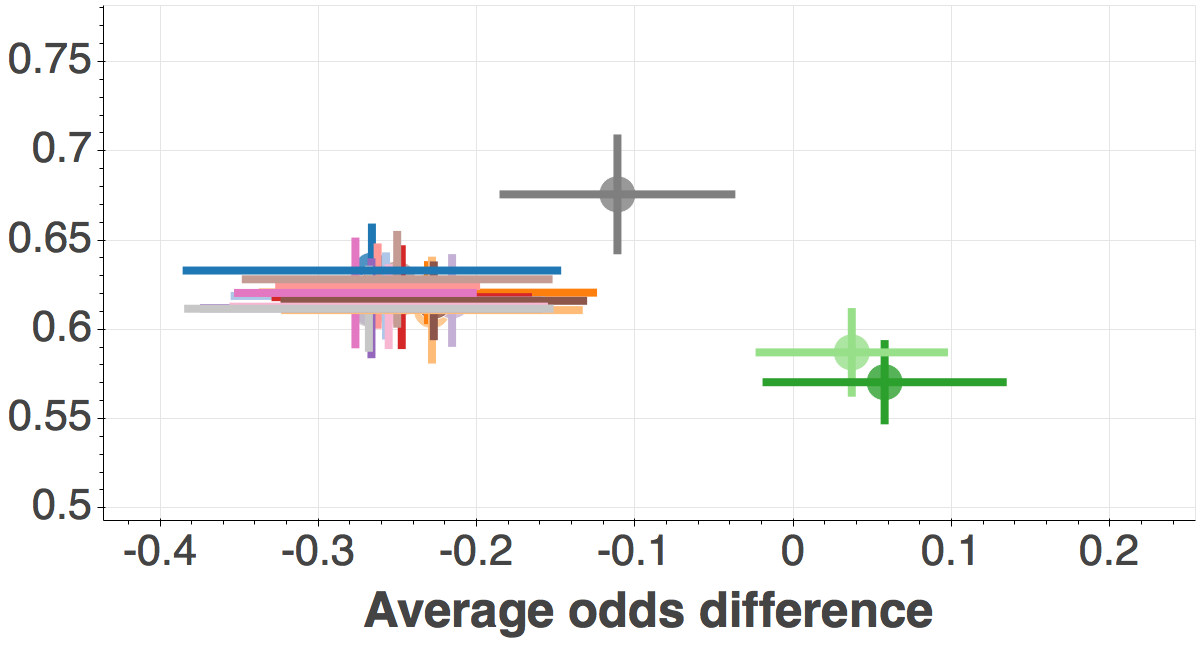}
    \includegraphics[scale=0.1]{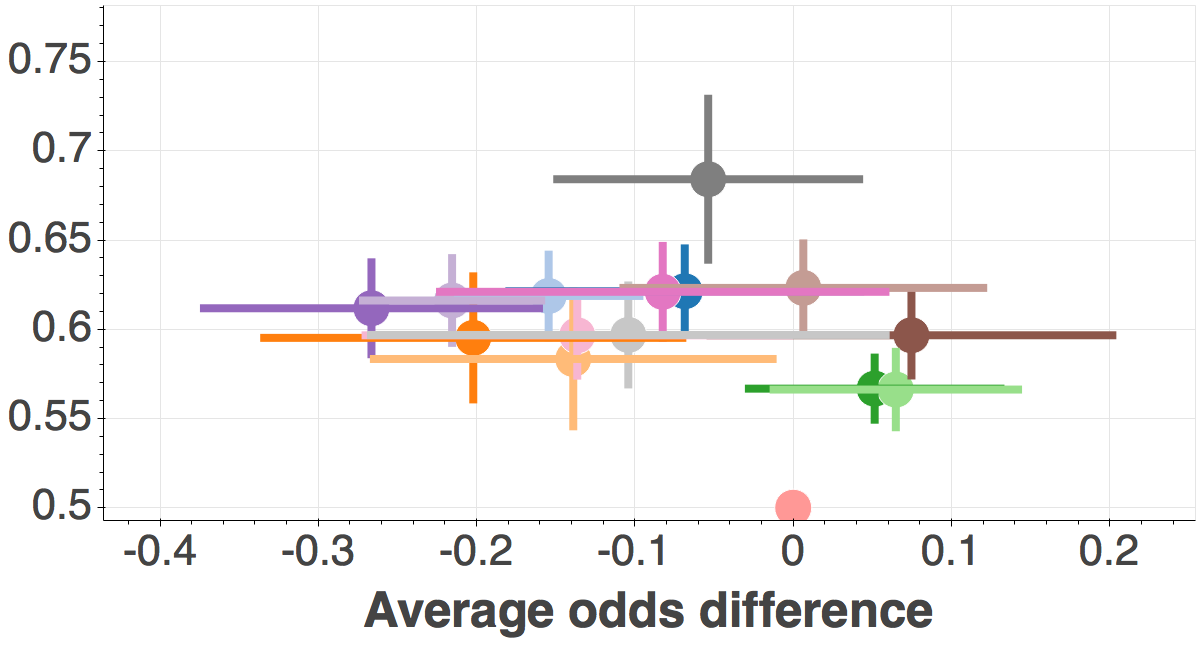}
    \label{fig:german_age_average_odds_difference}
    \caption{Average odds difference}
\end{subfigure}
\begin{subfigure}{0.245\textwidth}
    \centering
    \includegraphics[scale=0.1]{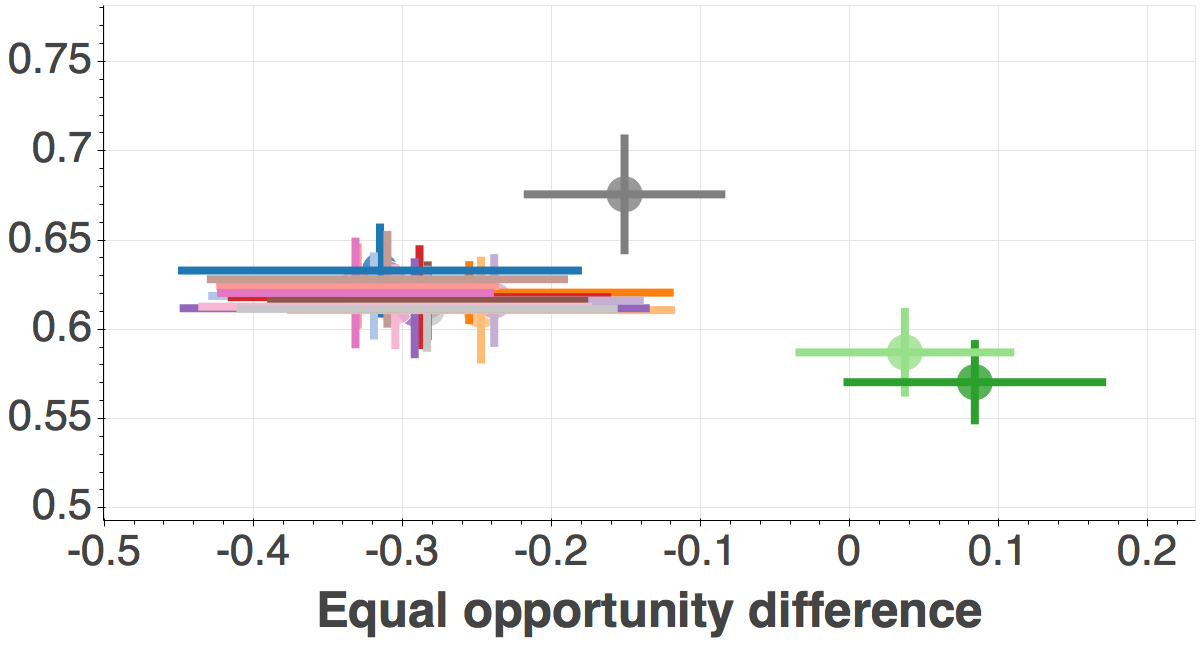}
    \includegraphics[scale=0.1]{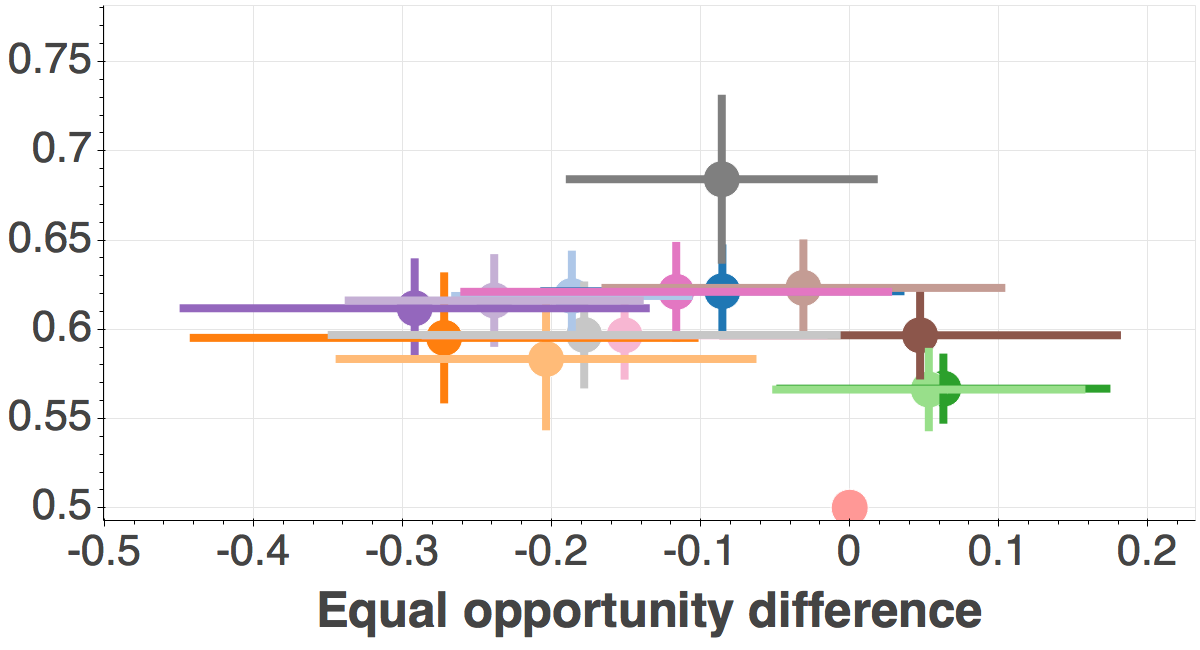}
    \label{fig:german_age_equal_opportunity_difference}
    \caption{Equal opportunity difference}
\end{subfigure}
\caption{Fairness vs. Balanced Accuracy before (top panel) and after (bottom panel) applying various bias mitigation algorithms. Four different fairness metrics are shown. In most cases two classifiers (Logistic regression - LR or Random forest classifier - RF) were used. The ideal fair value of disparate impact is 1, whereas for all other metrics it is 0. The circles indicate the mean value and bars indicate the extent of $\pm$1 standard deviation. Data set: \textit{german}, Protected attribute: \textit{age}. }
\label{fig:german-age}
\end{scriptsize}
\end{figure*}


\begin{figure*}[h!]
\begin{scriptsize}
\begin{subfigure}{1.0\textwidth}
    \centering
    \includegraphics[scale=0.25]{legend_hor.png}
\end{subfigure}
\begin{subfigure}{0.245\textwidth}
    \centering
    \includegraphics[scale=0.1]{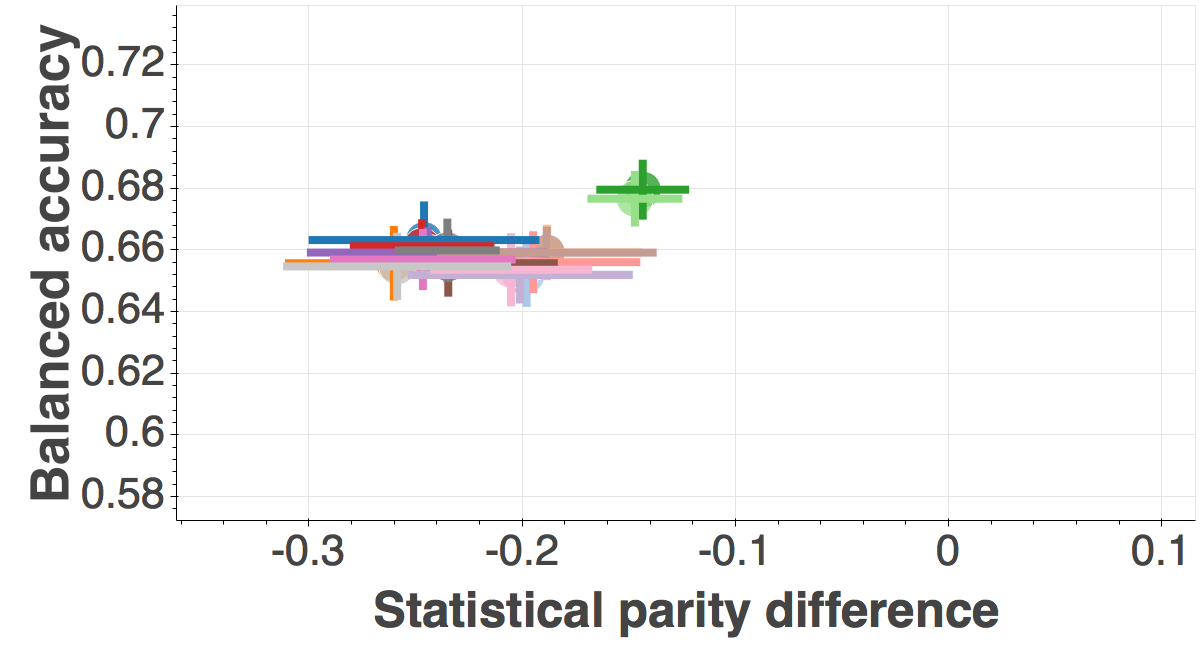}
    \includegraphics[scale=0.1]{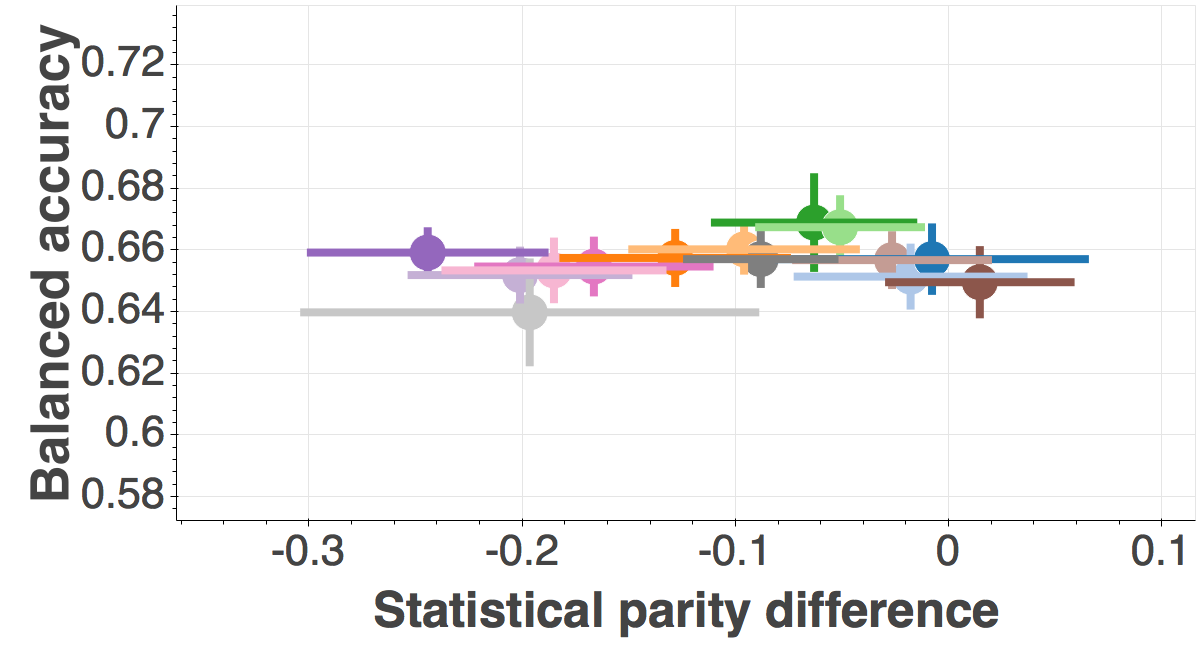}
    \label{fig:compas_sex_statistical_parity_difference}
    \caption{Statistical parity difference}
\end{subfigure}
\begin{subfigure}{0.245\textwidth}
    \centering
    \includegraphics[scale=0.1]{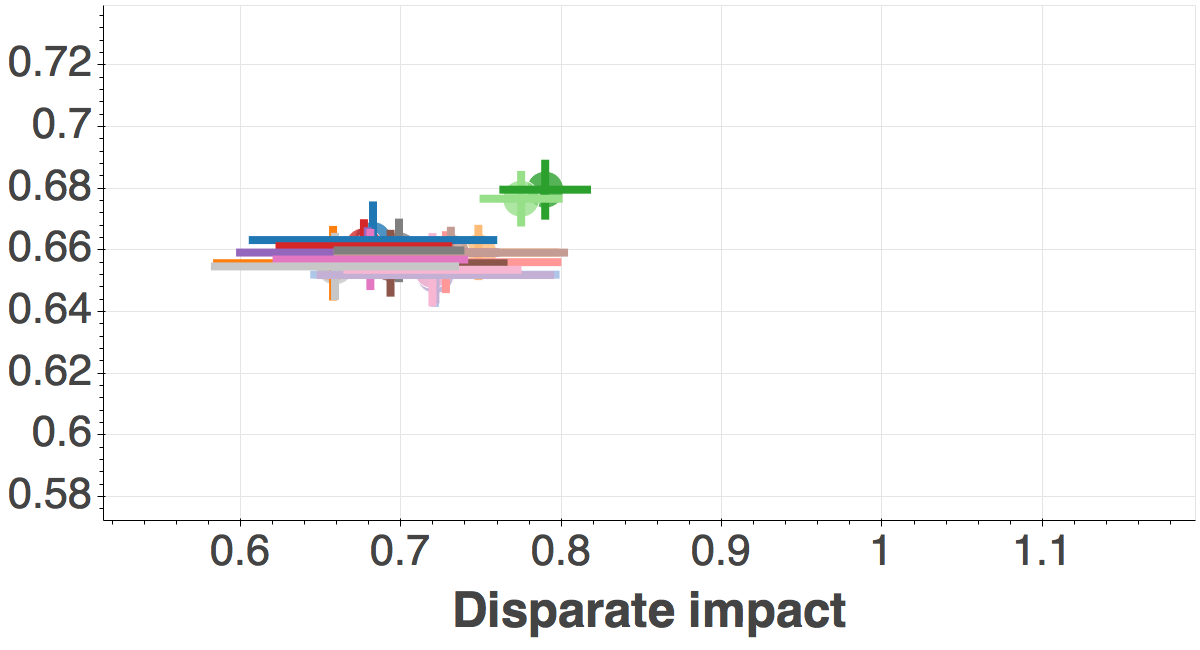}
    \includegraphics[scale=0.1]{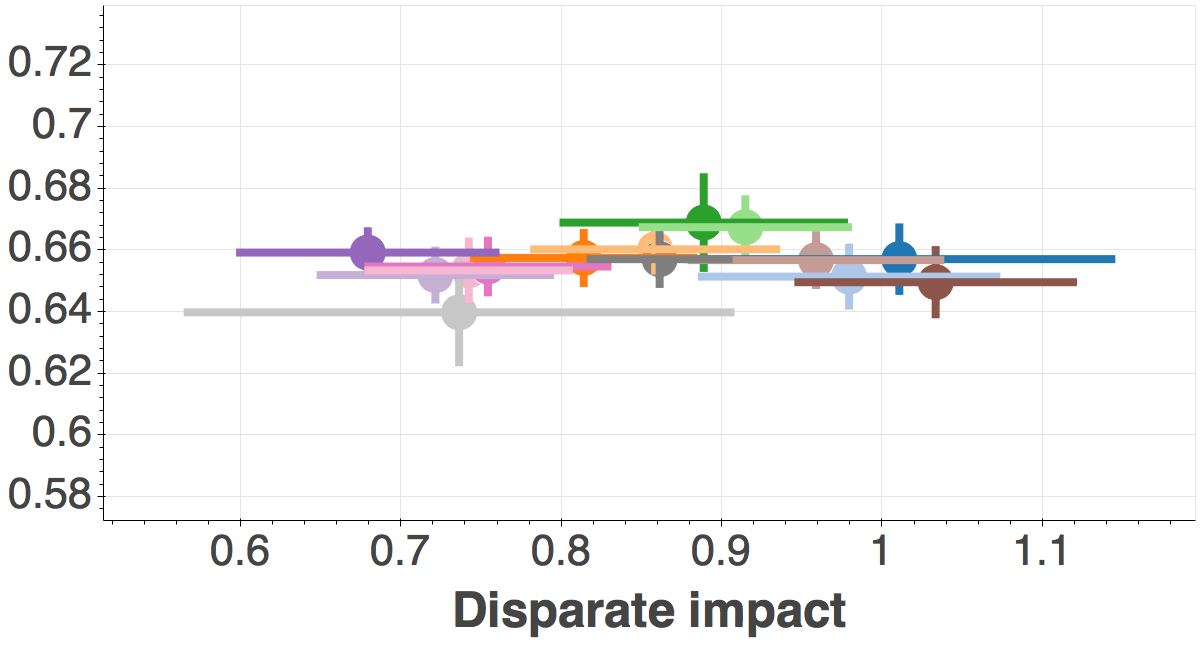}
    \label{fig:compas_sex_disparate_impact}
    \caption{Disparate impact}
\end{subfigure}
\begin{subfigure}{0.245\textwidth}
    \centering
    \includegraphics[scale=0.1]{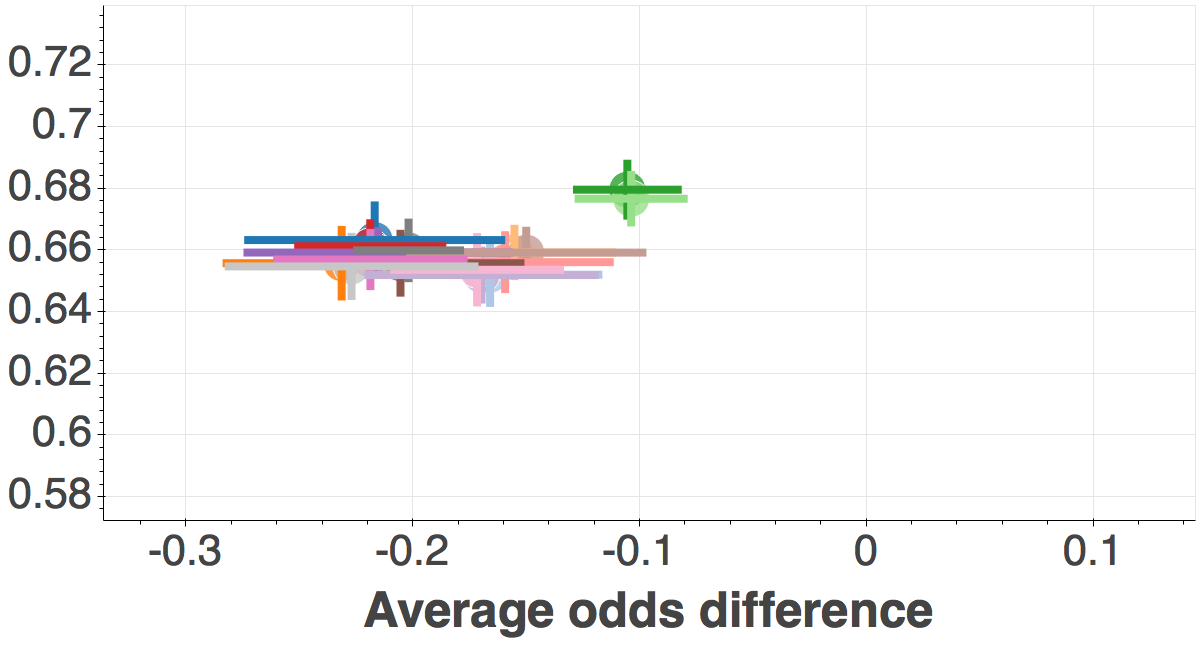}
    \includegraphics[scale=0.1]{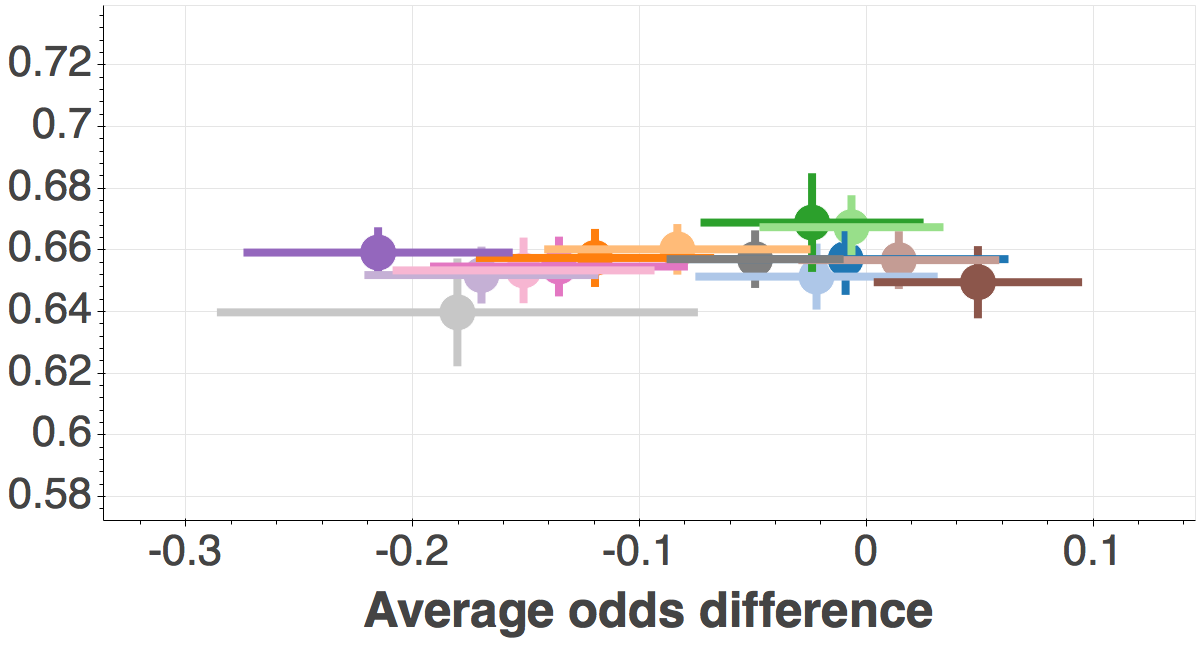}
    \label{fig:compas_sex_average_odds_difference}
    \caption{Average odds difference}
\end{subfigure}
\begin{subfigure}{0.245\textwidth}
    \centering
    \includegraphics[scale=0.1]{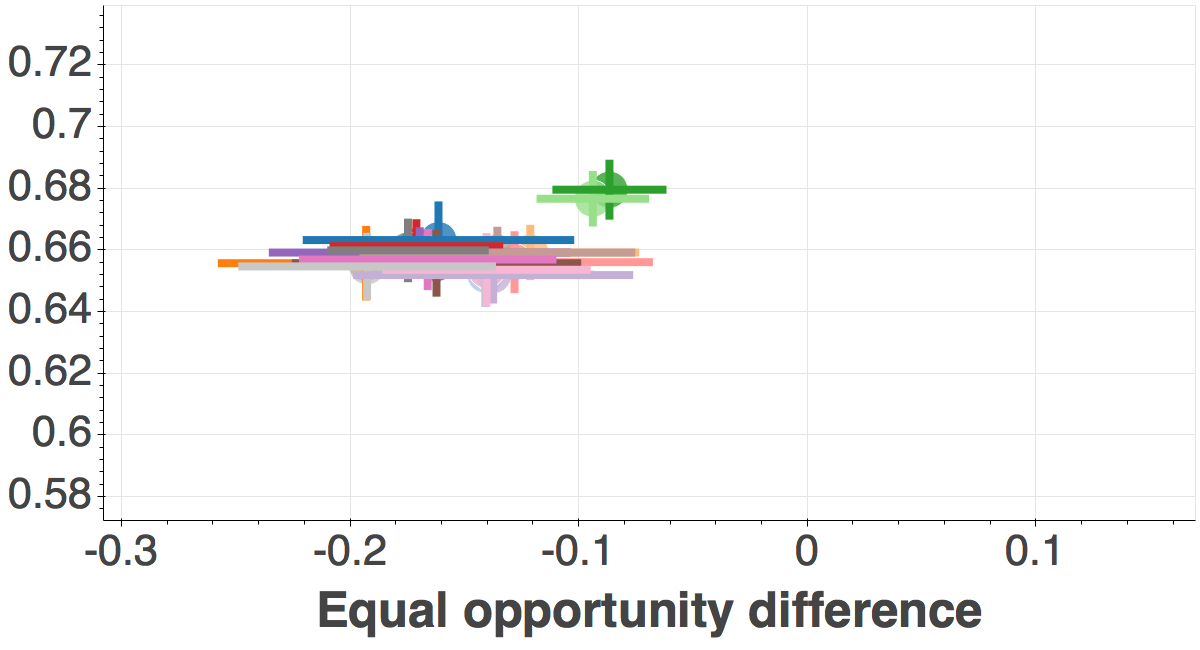}
    \includegraphics[scale=0.1]{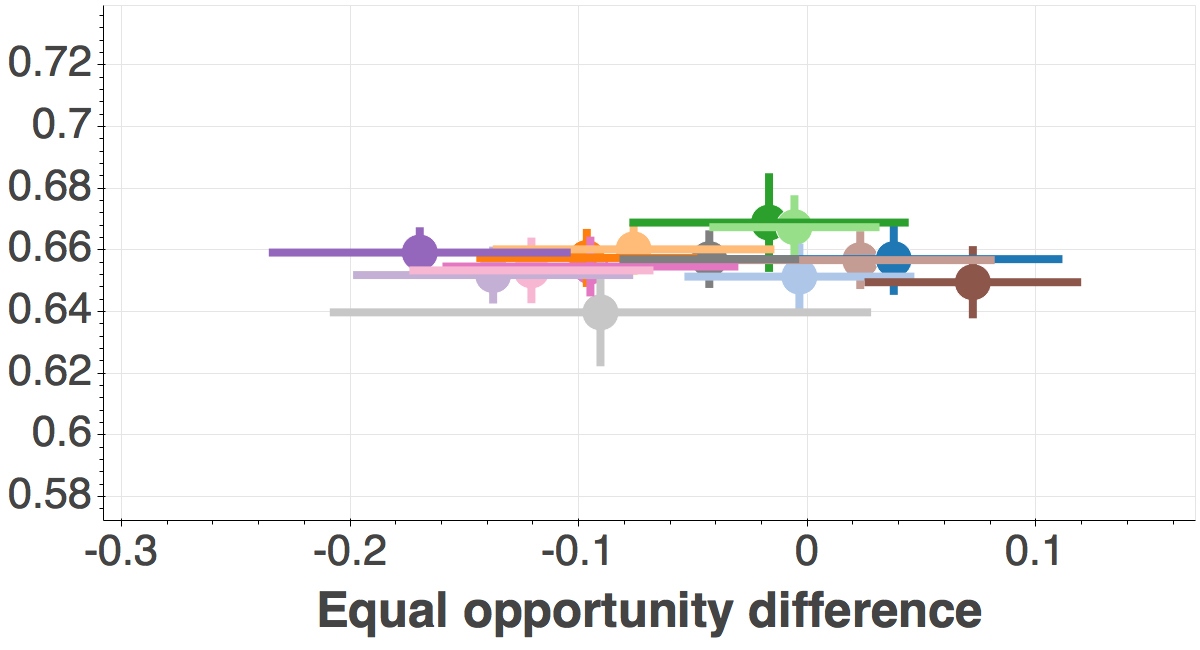}
    \label{fig:compas_sex_equal_opportunity_difference}
    \caption{Equal opportunity difference}
\end{subfigure}
\caption{Fairness vs. Balanced Accuracy before (top panel) and after (bottom panel) applying various bias mitigation algorithms. Four different fairness metrics are shown. In most cases two classifiers (Logistic regression - LR or Random forest classifier - RF) were used. The ideal fair value of disparate impact is 1, whereas for all other metrics it is 0. The circles indicate the mean value and bars indicate the extent of $\pm$1 standard deviation. Data set: \textit{compas}, Protected attribute: \textit{sex}. }
\label{fig:compas-sex}
\end{scriptsize}
\end{figure*}


\begin{figure*}[h!]
\begin{scriptsize}
\begin{subfigure}{1.0\textwidth}
    \centering
    \includegraphics[scale=0.25]{legend_hor.png}
\end{subfigure}
\begin{subfigure}{0.245\textwidth}
    \centering
    \includegraphics[scale=0.1]{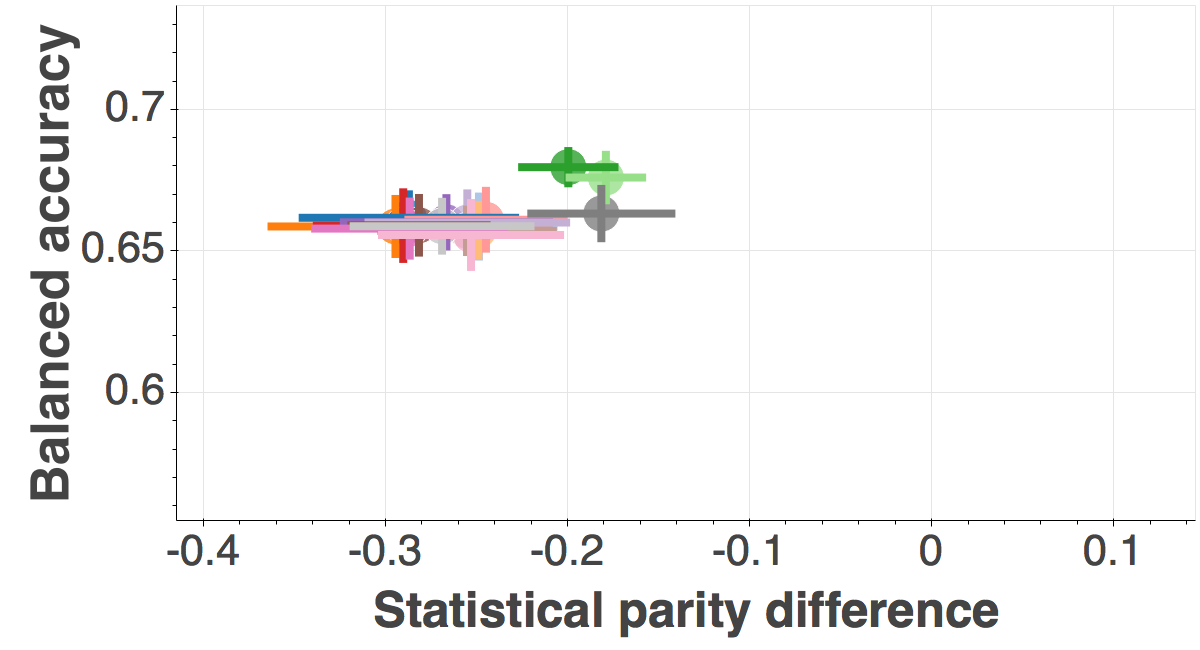}
    \includegraphics[scale=0.1]{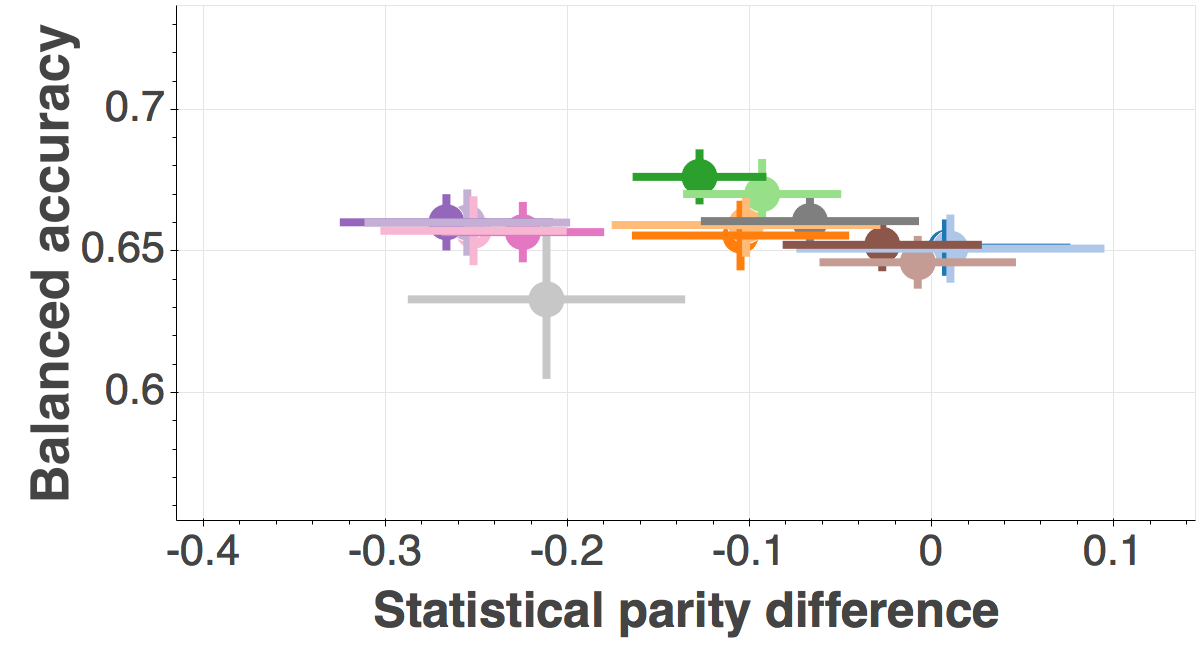}
    \label{fig:compas_race_statistical_parity_difference}
    \caption{Statistical parity difference}
\end{subfigure}
\begin{subfigure}{0.245\textwidth}
    \centering
    \includegraphics[scale=0.1]{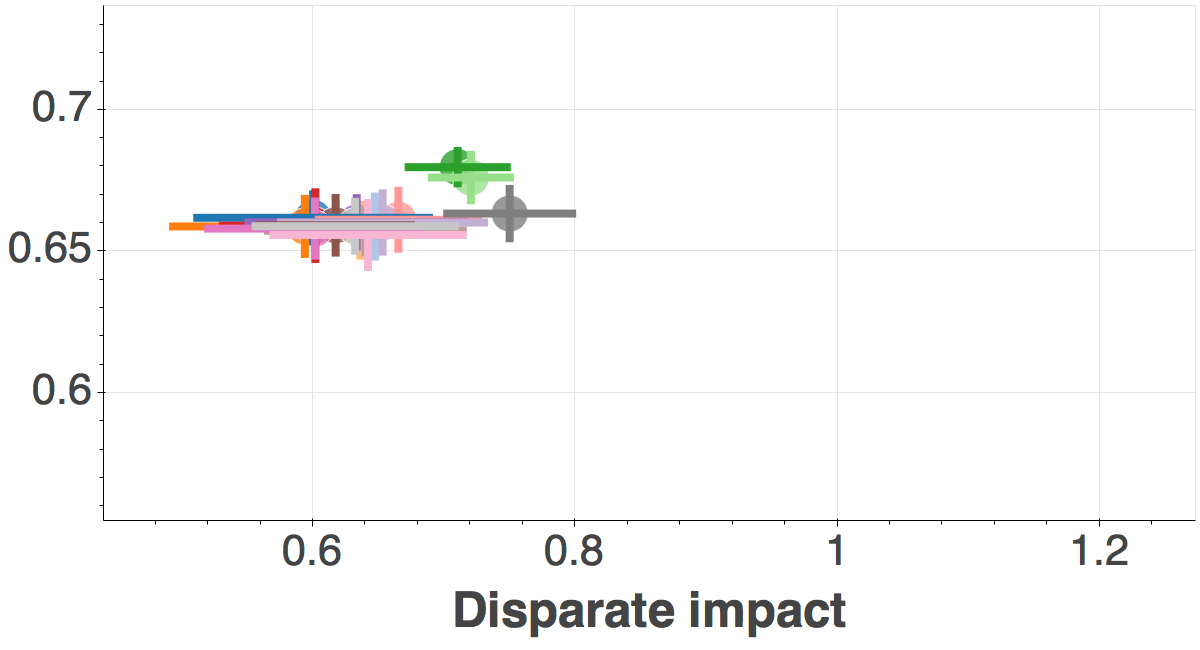}
    \includegraphics[scale=0.1]{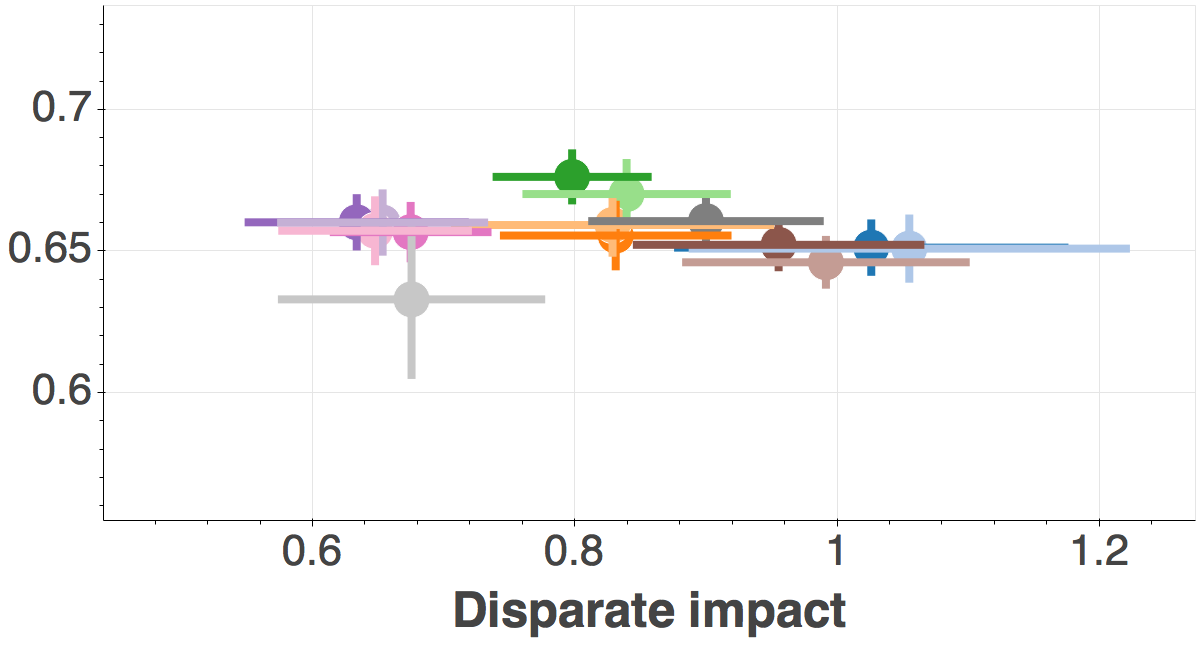}
    \label{fig:compas_race_disparate_impact}
    \caption{Disparate impact}
\end{subfigure}
\begin{subfigure}{0.245\textwidth}
    \centering
    \includegraphics[scale=0.1]{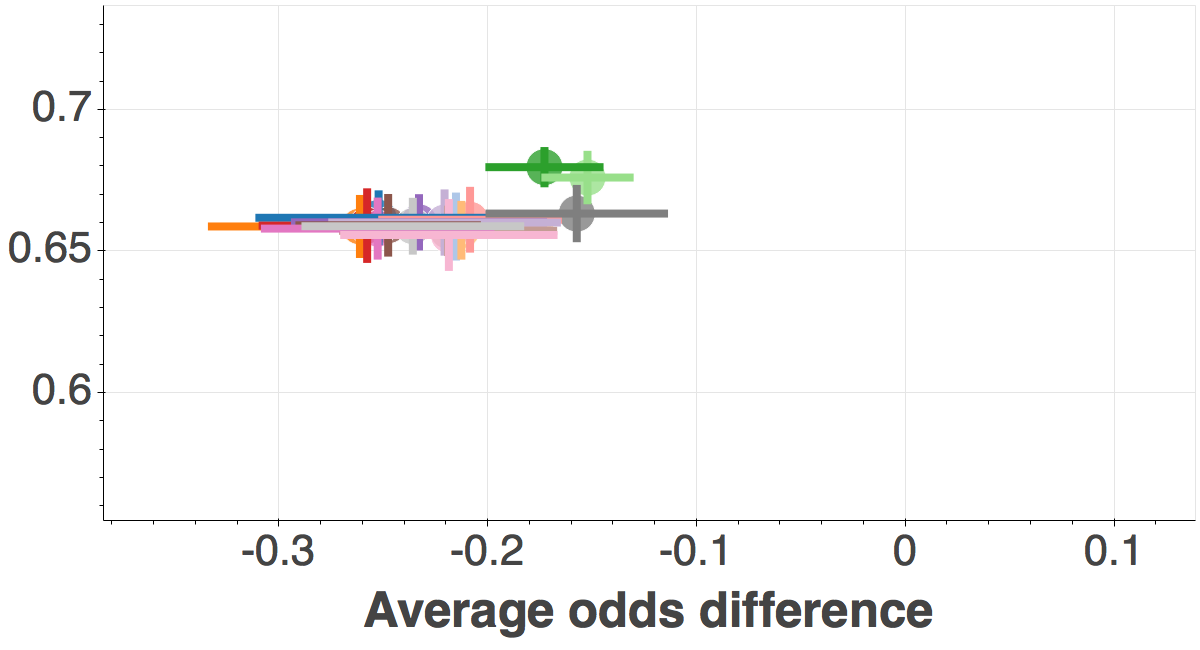}
    \includegraphics[scale=0.1]{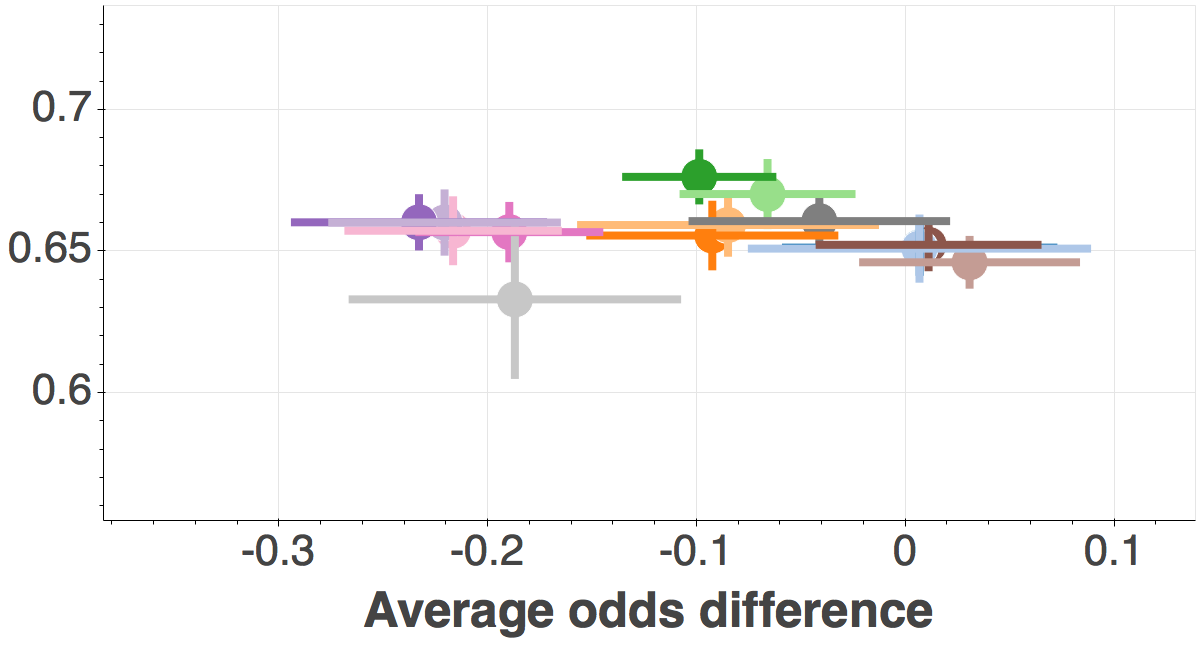}
    \label{fig:compas_race_average_odds_difference}
    \caption{Average odds difference}
\end{subfigure}
\begin{subfigure}{0.245\textwidth}
    \centering
    \includegraphics[scale=0.1]{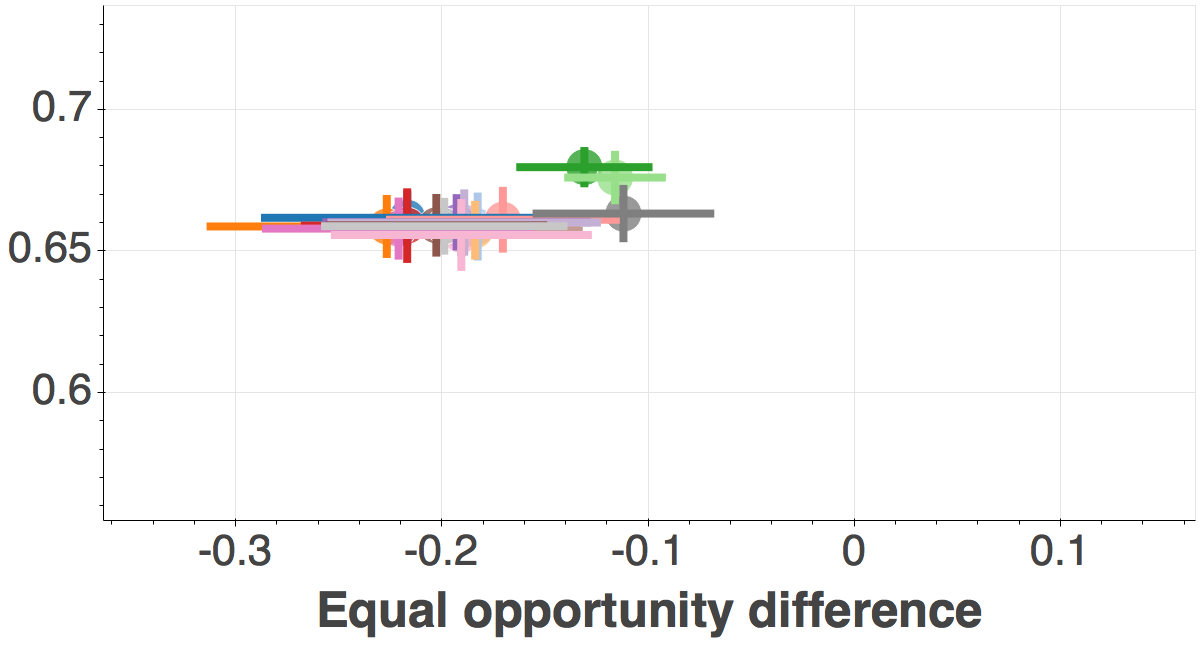}
    \includegraphics[scale=0.1]{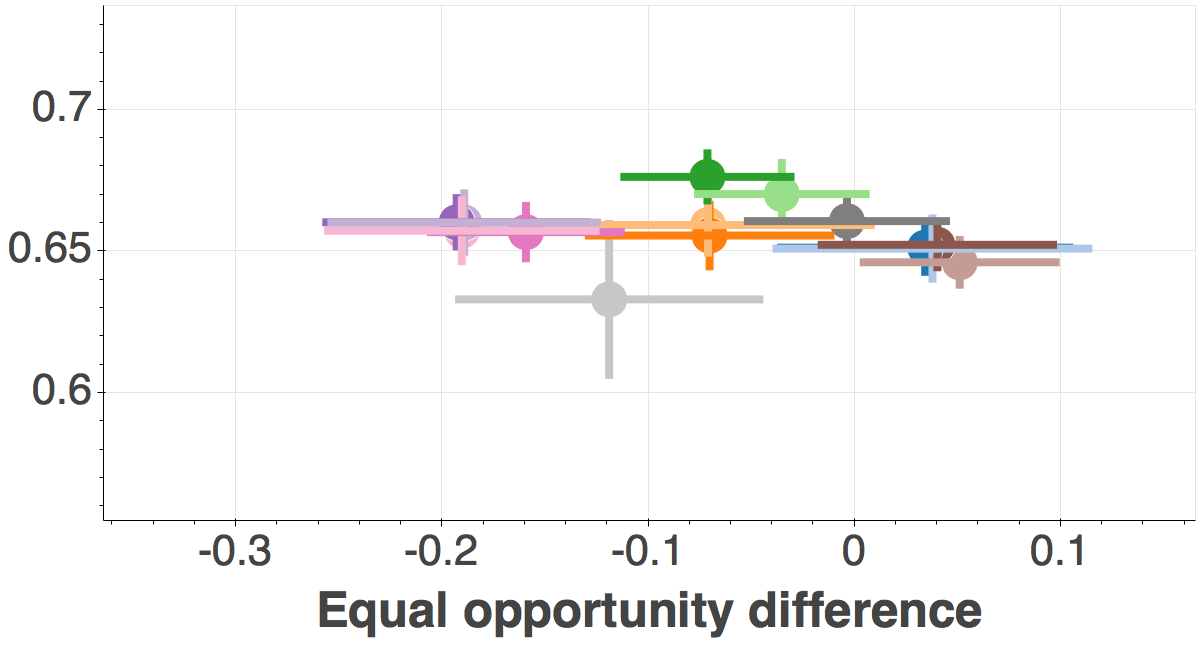}
    \label{fig:compas_race_equal_opportunity_difference}
    \caption{Equal opportunity difference}
\end{subfigure}
\caption{Fairness vs. Balanced Accuracy before (top panel) and after (bottom panel) applying various bias mitigation algorithms. Four different fairness metrics are shown. In most cases two classifiers (Logistic regression - LR or Random forest classifier - RF) were used. The ideal fair value of disparate impact is 1, whereas for all other metrics it is 0. The circles indicate the mean value and bars indicate the extent of $\pm$1 standard deviation. Data set: \textit{compas}, Protected attribute: \textit{race}. }
\label{fig:compas-race}
\end{scriptsize}
\end{figure*}


\clearpage

\section{UI page}
\label{appendix:posterpage}

\begin{figure*}[h]
  \centering
  \includegraphics[width=0.95\textwidth]{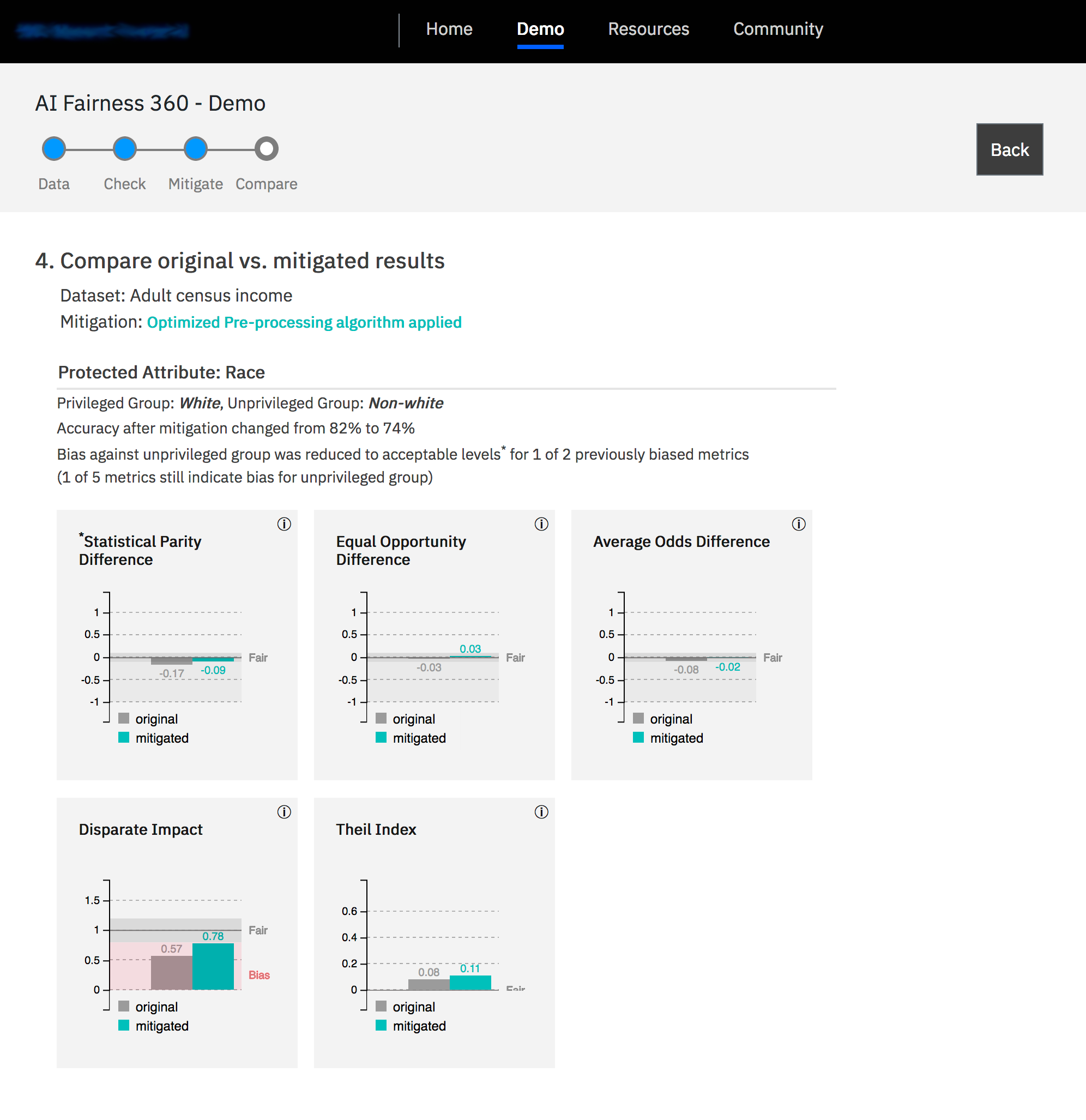}
  \caption{A screen shot from the web interactive experience, showing the results of mitigation applied to one of the available datasets.}
  \label{fig:posterpage}
\end{figure*}

\end{document}